%% file: main.tex
\documentclass{article}\PassOptionsToPackage{numbers,sort&compress}{natbib}
\usepackage[final]{colm2026_conference}\setcitestyle{numbers,square,comma}
\usepackage[utf8]{inputenc}
\usepackage[T1]{fontenc}
\usepackage{microtype}
\usepackage{url}
\usepackage{amsmath}
\usepackage{amsfonts}
\usepackage{amssymb}
\usepackage{mathtools}
\usepackage{amsthm}
\usepackage{booktabs}
\usepackage{colortbl}
\usepackage{graphicx}
\usepackage{subcaption}
\usepackage{float}
\usepackage{pifont}
\usepackage{xcolor}
\newcommand{\newprose}[1]{#1}
\newcommand{\vfiveprose}[1]{#1}
\newcommand{\purpleprose}[1]{#1}
\newcommand{\orangeprose}[1]{#1}
\long\def\redprose#1{#1}
\usepackage{tikz}
\usetikzlibrary{positioning, arrows.meta}
\usepackage{tcolorbox}
\tcbuselibrary{skins,breakable}
\usepackage{hyperref}
\usepackage{algorithm}
\usepackage{algorithmic}
\definecolor{softblue}{rgb}{0.88, 0.92, 1.0}
\definecolor{softgreen}{rgb}{0.88, 1.0, 0.88}
\definecolor{softred}{rgb}{1.0, 0.88, 0.88}
\newcommand{\pcorrect}{P(\text{correct})}
\newcommand{\cmark}{{\color{green!70!black}\ding{51}}}
\newcommand{\xmark}{{\color{red!70!black}\ding{55}}}
\definecolor{takeawayblue}{RGB}{53,90,150}
\definecolor{takeawayback}{RGB}{242,247,255}\newtcolorbox{takeaway}{  enhanced, breakable,  colback=takeawayback, colframe=takeawayblue,  boxrule=0.8pt, arc=3pt,  left=12pt, right=12pt, top=13pt, bottom=10pt,  before skip=1em, after skip=1em,  fontupper=\small,  title=Takeaway,  attach boxed title to top left={xshift=12pt, yshift=-8pt},  boxed title style={colback=takeawayblue, colframe=takeawayblue, boxrule=0pt, arc=1pt, left=8pt, right=8pt, top=3pt, bottom=3pt},  fonttitle=\bfseries\color{white}}
\definecolor{practgreen}{RGB}{184,110,0}
\definecolor{practback}{RGB}{255,248,230}\newtcolorbox{practnote}{  enhanced, breakable,  colback=practback, colframe=practgreen,  boxrule=0.8pt, arc=3pt,  left=12pt, right=12pt, top=13pt, bottom=10pt,  before skip=1em, after skip=1em,  fontupper=\small,  title=Disclaimer! \textsc{Pinocchio} may be miscalibrated on new data distributions,  attach boxed title to top left={xshift=12pt, yshift=-8pt},  boxed title style={colback=practgreen, colframe=practgreen, boxrule=0pt, arc=1pt, left=8pt, right=8pt, top=3pt, bottom=3pt},  fonttitle=\bfseries\color{white}}
\usepackage{lineno}\theoremstyle{plain}\theoremstyle{definition}\theoremstyle{remark}
\begin{document}
\title{Pinocchio: Fast Uncertainty Estimates for Black-Box Language Models}
\author{Kevin David Hayes\textsuperscript{1}, \ Arka Pal\textsuperscript{2}, \ Haosong Zhang\textsuperscript{3}, \ Tom Goldstein\textsuperscript{1}, \ Micah Goldblum\textsuperscript{2,4} \\
[0.4em]\textsuperscript{1}University of Maryland \quad \textsuperscript{2}Ritual AI \quad \textsuperscript{3}Fudan University \quad \textsuperscript{4}Columbia University}\ifcolmsubmission\linenumbers\fi
\maketitle{\let\thefootnote\relax\footnotetext{\textsuperscript{1}Department of Computer Science, University of Maryland. \textsuperscript{2}Ritual AI. \textsuperscript{3}Fudan University. \textsuperscript{4}Department of Computer Science and Department of Electrical Engineering, Columbia University. Correspondence: \texttt{khayes1@umd.edu}, \texttt{micah.g@columbia.edu}.}}
\input{sections/abstract}
\input{sections/introduction}
\input{sections/related_work}
\input{sections/method}
\input{sections/experiments}
\input{sections/discussion}
\clearpage
\bibliographystyle{plainnat}
\bibliography{refs_corrected_com}
\newpage
\appendix
\input{sections/appendix}
\end{document}

%% file: sections/abstract.tex
\begin{abstract}In high-stakes decision-making applications of large language models (LLMs), practitioners require not only accurate LLMs but also uncertainty estimates for their predictions. Existing approaches to uncertainty estimation for LLMs require access to log-probabilities output by the model or require fine-tuning access. However, many industrial LLM products use closed-source API models, and many such API models like GPT do not return log-probabilities and may not allow fine-tuning. We introduce \textsc{Pinocchio}, an external calibrator that estimates the correctness of responses from black-box API models. Trained jointly on responses from seven LLMs, it achieves 0.862 AUROC predicting the correctness of held-out responses from those same models, and shows zero-shot transfer to thirteen unseen models across eight organizations. Our model needs only a single forward pass to generate an uncertainty estimate and requires no access to the target model's logits, weights, or internal states. A lightweight text only 0.8B checkpoint matches our largest model's AUROC. We release code for adding uncertainty estimation to existing repos in only two additional lines of code.
\end{abstract}

%% file: sections/introduction.tex
\section{Introduction}
\label{sec:introduction}A physician asks GPT whether two medications interact. The model replies confidently, and incorrectly. Nothing in the API response flags this failure. This failure mode is not merely hypothetical, it is commonplace. As large language models are deployed in medicine, law, scientific research, and software engineering, wrong answers delivered with high confidence or without any indication of uncertainty at all remain an obstacle to trustworthy deployment. The closed-source models that dominate production (GPT-5, Claude, Gemini) expose little of the internal signal that might warn when they are wrong, and recent models increasingly hide even token log-probabilities, yet these are precisely the models that most need reliable uncertainty estimates. Existing approaches to uncertainty estimation for LLMs face a fundamental tension between access and performance. Logit- and representation-based methods~\citep{kadavath2022language} achieve reasonable performance but require white-box access that closed-source providers do not expose. Black-box alternatives like verbalized confidence~\citep{tian2023just, xiong2024llms}, self-evaluation~\citep{kadavath2022language}, and semantic entropy~\citep{kuhn2023semantic} pay a steep cost: verbalized confidence is poorly calibrated, self-evaluation inherits the model's blind spots, and semantic entropy requires 5--10$\times$ the inference budget. \textbf{Fast and accurate uncertainty estimation for black-box API models remains an open problem.}

\begin{figure}[t]\centering
\includegraphics[width=0.82\textwidth]{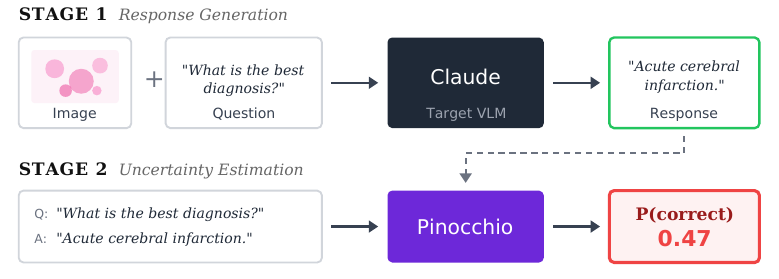}
\caption{\textbf{An external calibrator scores a target model's response without touching the model internals or logits.} It reads the question, response, and image and outputs a probability of correctness. This example illustrates the intended workflow; the displayed score $\pcorrect=0.47$ is illustrative. Adapted from MMMU~\citep{yue2024mmmu}.}
\label{fig:overview}
\end{figure}

We introduce \textsc{Pinocchio}, a family of learned uncertainty estimators that predict whether a model's response is correct from a single (question, response) pair, without access to log-probabilities, multiple samples, or model internals. Rather than extracting uncertainty from the target model itself, we train a small \emph{calibrator} (Figure~\ref{fig:overview}) that takes the question, the response, and the identity of the model that produced it, plus any associated images for vision-language tasks, and outputs a probability of correctness.
We train \textsc{Pinocchio} jointly on responses from seven LLMs that span a wide capability range (Section~\ref{sec:data}). It reaches 0.862 AUROC predicting the correctness of held-out responses, and \textsc{Pinocchio} also transfers zero-shot to thirteen models absent from training with a mean AUROC of 0.814 (Section~\ref{sec:cross_model}). A single forward pass produces a correctness probability without sampling, logit access, or modification of the target model. All of our uncertainty estimators are trained via supervised learning; the zero-shot transfer results are simply evaluations of the trained calibrator on LLMs whose responses it never saw, not a separate method. We summarize our contributions as follows:
\begin{enumerate}
\item \textbf{We release \textsc{Pinocchio}, an auxiliary uncertainty estimator for black-box LLMs. \textsc{Pinocchio}} is trained on responses from an ensemble of popular LLMs, including closed API models, reaching 0.862 AUROC on held-out responses with strong calibration relative to uncertainty-estimation baselines.  Estimating uncertainty requires only a single forward pass, without access to log-probabilities, sampling, or model weights, and assigns correctness probabilities to responses from models that expose no internal signal.
\item \textbf{A small uncertainty estimator.} Our released 0.8B checkpoint retains 99\% of our largest model's AUROC.
\item \textbf{Zero-shot transfer across model families.} The same calibrator transfers to thirteen unseen models across eight organizations.
\item \textbf{Vision-language data supplies training signal that transfers to text.} \vfiveprose{Strong} models saturate many text-only datasets, so errors are scarce. Vision-language benchmarks still elicit frequent errors, and what the estimator learns from them transfers to text-only benchmarks, improving text-only calibration when hard text examples are limited (Section~\ref{sec:per_benchmark}). 
\end{enumerate}

%% file: sections/related_work.tex
\section{Related work}
\label{sec:related_work}
\paragraph{Uncertainty quantification in LLMs.}Existing methods fall into three categories. \emph{Token-level} approaches use output probabilities \citep{kadavath2022language, plaut2024softmax} but conflate confidence over a sequence of tokens with confidence over the answer non-uniquely expressed by that sequence. \emph{Sampling-based} methods measure consistency across multiple generations \citep{kuhn2023semantic, farquhar2024semanticentropy, chenmueller2024bsdetector, hamidieh2026crossmodel}, but require a large number of forward passes, multiplying API costs. \emph{Verbalized} approaches prompt models to state their confidence \citep{lin2022teaching, tian2023just, mielke2022reducing}, but LLMs are systematically overconfident \citep{xiong2024llms, groot2024overconfidence, heo2025llmuq}. When model internals are accessible, probing hidden states \citep{burns2022latentknowledge, azaria2023internalstate, kossen2024sep} or training supervised UQ modules \citep{shelmanov2025uqheads, khanmohammadi2025ccps} yields stronger signals, but these white-box methods are inapplicable to closed-source models.  Selective prediction and generation instead use confidence to abstain or control prediction and generation~\citep{geifman2017selective, lee2024selectivegeneration}.

\paragraph{Fine-tuned uncertainty judges.} Fine-tuning can teach an answering model to expose or act on its uncertainty. \citet{lin2022teaching} supervise GPT-3 to generate calibrated verbal confidence, including under distribution shift. R-Tuning~\citep{zhang2024rtuning} constructs refusal-aware instruction data so that a model learns to answer known questions and abstain on questions outside its knowledge. Both approaches require training access to the model whose uncertainty they seek to improve. Relevant to our work, \citet{kapoor2024llms} train probes and LoRA adapters on the target model's hidden states to predict the target model's response correctness, outperforming black-box baselines but requiring open-weight access to the target model. APRICOT~\citep{ulmer-etal-2024-calibrating} is the most closely related black-box approach to \textsc{Pinocchio}: this work trains an auxiliary model to predict an LLM's confidence from input-output pairs, unlocking uncertainty estimation for black-box API models. \textsc{Pinocchio} extends this work by training a single calibrator on multiple target models simultaneously, introducing model-identity tags, and demonstrating the benefit of training on multiple target models. We further show that \textsc{Pinocchio} generalizes to unseen target models, handles vision-language problems, and we use newer and stronger target models that saturate many benchmarks and require carefully constructing a training set for \textsc{Pinocchio} that contains sufficiently many examples of the target models being incorrect.

\paragraph{Data scaling and vision-language uncertainty.} Table~\ref{tab:literature} summarizes how existing UQ methods meet the desiderata we target. Non-learned methods (verbalized confidence, semantic entropy, LLM-as-judge) have no mechanism to improve with additional data. In an ablation, \textsc{Pinocchio}'s performance scales log-linearly with training examples
(Section~\ref{sec:ablations}). Strong LLMs still err frequently on vision-language benchmarks \citep{li2023pope, guan2023hallusionbench}, which makes them a rich source of the hard training examples a calibrator needs; uncertainty estimation for VLMs itself remains underexplored.  \textsc{Pinocchio} treats these frequent vision-language errors as a rich training signal, and that signal transfers to text, so a single checkpoint trained jointly to estimate uncertainty on vision-language and language-only tasks exhibits strong performance on both.  Existing uncertainty estimation methods each drop at least one property we require: they need multiple forward passes \citep{zhang2024vluncertainty}, depend on grounding annotations \citep{padhi2025grounding}, or train a critic bound to one target model \citep{zhang2025criticv}.  Others \citep{zhang2025uncertaintyo, kostumov2024vlm} have been evaluated only within a single model family, leaving cross-model-family generalization untested. To our knowledge, no prior work delivers a black-box, single-pass uncertainty estimator that learns from data and generalizes across models.
\begin{table}[t]\centering
\scriptsize
\caption{\textbf{\textsc{Pinocchio} estimates uncertainty for black-box API models in one forward pass, improves with data, transfers across model families, and handles multimodal targets.}}
\label{tab:literature}
\begin{tabular}{@{}lccccc@{}}\toprule
Method & Black-box & Single-pass & \shortstack{Scales\\ with data} & \shortstack{Target model \\ agnostic} & Multimodal \\
\midrule
Logit-based \citep{kadavath2022language} & \xmark & \cmark & \xmark & \vfiveprose{\cmark} & \xmark \\
MC Dropout \citep{gal2016dropout} & \xmark & \xmark & \xmark & \vfiveprose{\cmark} & \xmark \\
Semantic Entropy \citep{kuhn2023semantic} & \cmark & \xmark & \xmark & \vfiveprose{\cmark} & \xmark \\
Verbalized Conf. \citep{tian2023just} & \cmark & \cmark & \xmark & \vfiveprose{\cmark} & \xmark \\
ORM / Verifiers \citep{cobbe2021training} & \xmark & \cmark & \cmark & \xmark & \xmark \\
LLM-as-Judge \citep{zheng2023judging} & \cmark & \cmark & \xmark & \vfiveprose{\cmark} & \xmark \\
White-box Probe \citep{kapoor2024llms} & \xmark & \cmark & \cmark & \xmark & \xmark \\
APRICOT \citep{ulmer-etal-2024-calibrating} & \cmark & \cmark & \cmark & \xmark & \xmark \\
\midrule
\textsc{Pinocchio} (ours) & \cmark & \cmark & \cmark & \cmark & \cmark \\
\bottomrule
\end{tabular}

\end{table}

%% file: sections/method.tex
\section{Method}
\label{sec:method}
\subsection{Problem formulation}
\label{sec:problem} Let $Q$ denote a question, including any images associated with it, and let $A$ be a candidate answer produced by target model $M_{\text{target}}$, including any chain-of-thought or reasoning tokens the API exposes. We train $f(Q,A,M) \in [0,1]$ to infer whether $A$ is correct, where $M$ is an optional model-identity tag. The training set can then be written $\mathcal{D}=\{(Q_i,A_i,M_i,y_i)\}_{i=1}^N$, with $y_i\in\{0,1\}$.  \textsc{Pinocchio} outputs $P(y=1\mid Q,A,M)$. We allow the calibrator to condition on the model-identity tag $M$ so that it can use model-specific patterns to identify uncertainty, noting a slight gain over model-agnostic estimation (Appendix~\ref{app:scaling}). When predicting uncertainty for new target models, we simply omit the model-identity tag. We measure both discrimination and calibration in Section~\ref{sec:experiments} using AUROC, Brier score, and ECE.
\subsection{Architecture and training}
\label{sec:architecture}We fine-tune Qwen3-VL-8B-Instruct~\citep{qwen2025qwen3vl} with LoRA~\citep{hu2021lora} (rank 32, $\alpha = 64$) on all linear layers. This VLM processes both text and image inputs in a single architecture. Text-only benchmarks have no images, so we fill the image slot with a blank uniform-gray placeholder that carries no visual signal, keeping a single input format for text and vision-language examples. We format each input as a chat-style prompt containing the question, response, and optional benchmark and model-identity metadata (Appendix~\ref{app:prompts}); including metadata raises AUROC from 0.813 to 0.863 (Section~\ref{sec:ablations}).  The model predicts a single token, \texttt{i} (incorrect) or \texttt{ii} (correct), and we extract the correctness probability via softmax over the two logits:
\begin{equation}    \pcorrect = \frac{\exp(z_{\text{ii}})}{\exp(z_{\text{i}}) + \exp(z_{\text{ii}})}
\end{equation}
We train the model with cross-entropy loss on this final token only. Full training configuration is in Appendix~\ref{app:training_config}; hyperparameters in Appendix~\ref{app:hyper}; training data breakdown in Appendix~\ref{app:data}.
\subsection{Training data}
\label{sec:data}Since strong models achieve very high scores on popular benchmarks, the central challenge in training a calibrator is assembling enough examples where strong models are wrong. We address this deficit by selecting benchmarks with both correct and incorrect responses, including vision-language data where models tend to be weaker, and pooling responses from several target models. Our training mixture contains responses from \textbf{Claude Fable~5}, \textbf{Claude Opus~5}, \textbf{GPT-5.6}, and \textbf{Kimi~3}, together with lower-capability sources \textbf{GPT-5-mini}, \textbf{GPT-5.2}, and \textbf{Qwen3.5-397B}, across 20 benchmarks. Refusals and empty responses are retained and labeled incorrect. None of the thirteen target models we use in zero-shot transfer experiments appear in this training mixture. We grade responses using exact or fuzzy matching on 15 benchmarks and GPT-5-mini as a judge on the five that require rubric-based grading; Appendix~\ref{app:self_preference} tests the robustness of the latter labels. All responses to the same question are assigned to the same split.
\paragraph{Benchmark selection.}We design the benchmark mixture around three criteria. First, \emph{difficulty calibration}: we prioritize benchmarks where strong models achieve 20--80\% accuracy, ensuring a sufficiently balanced class distribution. Benchmarks that are too easy (e.g., MMLU, $>$85\% accuracy) or too hard (e.g., FrontierMath, $<$5\%) provide little discriminative signal. Second, \emph{domain diversity}: we span seven domains (Table~\ref{tab:benchmarks}) to prevent the calibrator from learning domain-specific shortcuts: consistent with this, leave-$K$-out cross-validation shows a mean AUROC gap of only 0.002 between included and excluded benchmarks (Appendix Table~\ref{tab:leave_k_out}), i.e.\ no single domain is crucial. Third, \emph{format and modality heterogeneity}: we include multiple-choice, open-ended, and binary formats across both text-only and vision-language modalities, teaching the calibrator general correctness signals rather than format-specific heuristics.

\paragraph{Multi-model training.} Pooling responses from several target models increases both the amount and variety of training data. Successively adding lower-capability sources to the pool of high-capability sources improves mean transfer across eleven unseen models (Table~\ref{tab:capability_diversity}).
\begin{table}[t]\centering
\footnotesize
\caption{\textbf{The mixture spans 7 domains and 2 modalities in the 20--80\% accuracy band.} Benchmark mixture spanning 7 domains and 2 modalities. Of 20 benchmarks, 9 are text-only and 11 are vision-language.}
\label{tab:benchmarks}
\begin{tabular}{@{}llc@{}}\toprule
\textbf{Domain} & \textbf{Benchmarks} & \textbf{Modality} \\
\midrule
Mathematics & OmniMath, MathVista, MathVerse, MathVision & T+V \\
Science & GPQA Diamond, ChemBench & T  \\
Professional and coding & PRBench, ARC-AGI & T \\
General & SimpleQA, LiveBench, BBEH & T \\
Multimodal & MMMU, MMStar, MM-Vet, CharXiv, VizWiz & V \\
Spatial & RealWorldQA, HallusionBench & V \\
Frontier & HLE, HLE-Multimodal & T+V \\
\bottomrule
\end{tabular}

\end{table}

%% file: sections/experiments.tex
\section{Experiments}
\label{sec:experiments}
\subsection{Experimental setup}
\paragraph{Target models.} \textsc{Pinocchio} is trained jointly on responses from the seven models described in Section~\ref{sec:data}. We evaluate the same calibrator on held-out questions answered by those same models whose responses were used for training and zero-shot on thirteen target models absent from training (Section~\ref{sec:cross_model}). Unless otherwise stated, experiments use our 8B calibrator; Section~\ref{sec:ablations} evaluates smaller variants.
\paragraph{Benchmarks.}We evaluate on 20 benchmarks across 7 domains, including 9 text-only and 11 vision-language benchmarks. Full benchmark details and citations are in Appendix~\ref{app:benchmarks}.
\paragraph{Baselines.}
\label{sec:baselines}We compare against verbalized confidence, Platt and isotonic recalibration, response length, an LLM judge, and proxy semantic entropy and self-consistency computed from $N{=}5$ Qwen3-VL-8B samples. The first six use the full held-out test set of 1{,}953 samples (question-level split); the two proxy sampling baselines use its 1{,}376-example text-only subset. Because the proxy variants sample a stand-in model rather than the target, we additionally evaluate a full suite of \emph{faithful} sampling estimators run directly on open targets: semantic entropy~\citep{kuhn2023semantic}, self-consistency~\citep{wang2023selfconsistency}, the graph-theoretic measures of \citet{lin2023generating}, SPUQ~\citep{gao2024spuq}, and the logit-free conformal method of \citet{su2024api} (Section~\ref{sec:main_results}; Appendix~\ref{app:faithful_baselines}). Implementation details are found in Appendix~\ref{app:baselines}. \redprose{We additionally evaluate a purpose-built decision model, the commercial TypeSafe Jev, which does not provide a usable correctness signal off the shelf (Appendix~\ref{app:decision_models}).}
\paragraph{Metrics.}We report \textbf{AUROC} (area under the receiver operating characteristic curve), which measures the ability to rank correct responses above incorrect ones, regardless of threshold. We include 95\% BCa bootstrap confidence intervals from 2{,}000 resamples. We additionally report Brier score and expected calibration error (ECE; 15 equal-width bins)~\citep{guo2017calibration} for calibration, and use DeLong's test~\citep{delong1988comparing} for paired full-set AUROC comparisons.
\subsection{Main results}
\label{sec:main_results} 
\begin{figure}[t]\centering
\includegraphics[width=0.74\textwidth]{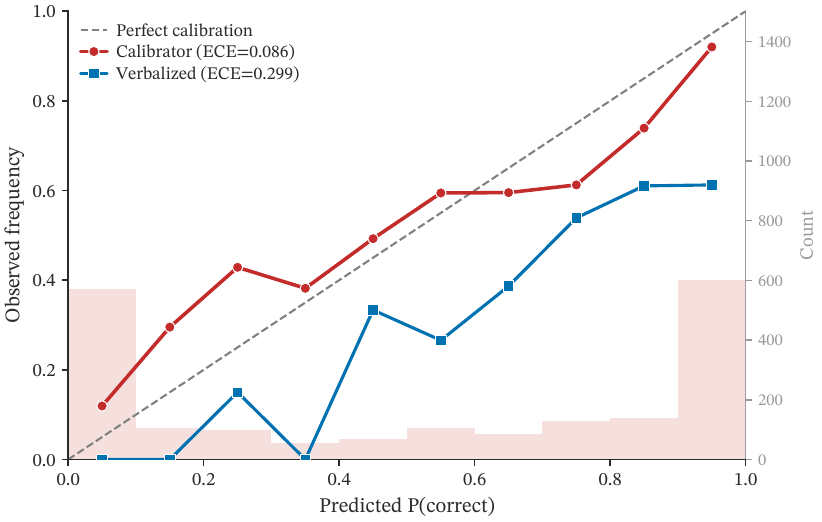}
\caption{\textbf{\textsc{Pinocchio}'s probabilities are well-calibrated; raw verbalized confidence is not.} Reliability diagram on the held-out test set. \textsc{Pinocchio}'s probabilities track observed accuracy far more closely than raw verbalized confidence, which remains overconfident on incorrect responses.}
\label{fig:calibration_main}
\end{figure}

\begin{figure}[t]\centering
\includegraphics[width=0.55\textwidth]{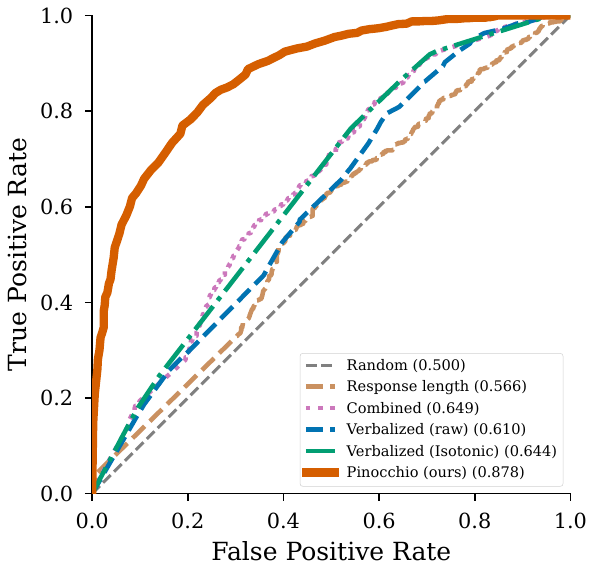}
\caption{\textbf{\textsc{Pinocchio} separates correct from incorrect responses far more cleanly than single-pass baselines.} ROC curves on the held-out test set of 1{,}953 examples.}
\label{fig:roc_main}
\end{figure}

\textsc{Pinocchio} separates correct from incorrect responses far better than any single-pass black-box baseline (0.86 vs.\ 0.65 AUROC) and is far better calibrated than a model's own verbalized confidence (Table~\ref{tab:main}). Per-model AUROC, Brier, and ECE are in Appendix Table~\ref{tab:frontier_flagship}.
\begin{table}[t]\centering
\footnotesize
\caption{\textbf{A trained calibrator predicts black-box correctness far better than any single-pass baseline.} Main results on the held-out test set (1{,}953 examples, question-level split). The \textsc{Pinocchio} row contains the released calibrator; baselines are computed on the same responses). Brackets give 95\% bootstrap CIs. Differences between \textsc{Pinocchio} and each full-set baseline are significant by DeLong's test (Appendix~\ref{app:effect_size}).}
\label{tab:main}
\begin{tabular}{@{}lccc@{}}\toprule
Method & AUROC $\uparrow$ & Brier $\downarrow$ & ECE $\downarrow$ \\
\midrule
Verbalized (raw) & 0.610 [.590, .624] & 0.326 & 0.299 \\
Verbalized (Platt) & 0.610 [.590, .624] & 0.232 & 0.034 \\
Verbalized (Isotonic) & 0.644 [.620, .667] & 0.231 & 0.009 \\
Response length & 0.566 [.542, .592] & 0.237 & 0.028 \\
Combined (verb.+len.) & 0.649 [.624, .673] & 0.229 & 0.021 \\
LLM-as-judge & 0.532 [.524, .540] & 0.257 & 0.110 \\
\redprose{TypeSafe Jev (zero-shot)$^*$} & \redprose{0.666 [.638, .691]} & \redprose{0.240} & \redprose{0.133} \\
Proxy sem.\ entropy$^\dagger$ & 0.467 & 0.340 & 0.266 \\
Proxy self-consist.$^\dagger$ & 0.466 & 0.340 & 0.266 \\
\midrule
\textsc{Pinocchio} (ours) & \textbf{0.863} [.847, .879] & \textbf{0.164} & 0.107 \\
\bottomrule
\end{tabular}
\par\vspace{2pt}{\scriptsize $^\dagger$Text-only subset (1{,}376 samples); CIs omitted. \redprose{$^*$Text-only subset.}}
\end{table}
Baseline and ablation comparisons use the matched 1{,}953-response test set. On this set, \textsc{Pinocchio} achieves \purpleprose{0.863} AUROC at predicting the target model's correctness, compared with 0.649 for the strongest single-pass black-box baseline (Table~\ref{tab:main}; Appendix~\ref{app:additional}). Figures~\ref{fig:calibration_main} and~\ref{fig:roc_main} use the same evaluation set. The single-pass and sampling baselines were computed once on this set and were not rerun for \textsc{Pinocchio} or newer transfer targets, several of which require per-target model or API access; calibration of \textsc{Pinocchio} is reported in Tables~\ref{tab:frontier_flagship} and~\ref{tab:recal_unseen}.

Appendix~\ref{app:effect_size} reports significance tests and how large each gap is for the baseline comparisons, and Appendix~\ref{app:scaling} reports further input ablations.
\begin{takeaway}A small external calibrator predicts black-box correctness far better than the confidence a model expresses about its own responses (0.88 vs.\ 0.65 AUROC). Our external calibrator is also far better calibrated than raw verbalized confidence.
\end{takeaway}
On the 397 test questions where the models whose responses we used to train \textsc{Pinocchio} disagree (at least one response correct and at least one incorrect, so every response to a question shares the same difficulty), \textsc{Pinocchio} still ranks the correct responses above the incorrect ones, which no predictor blind to the response could do (Appendix~\ref{app:disagreement}). This experiment shows that our model reads the response, not just the difficulty of the question.  An input ablation separates the two cues the calibrator could rely on: how hard the question is for the model and the content of the answer itself (Table~\ref{tab:knowledge_cues}). Removing the question leaves most of the signal intact, so the calibrator reads the answer, not just question difficulty; but the question still contributes, so it uses both.
\begin{takeaway}The calibrator relies on both identifying questions that are hard for the target model and interpreting linguistic cues in the target model's response.
\end{takeaway}
If the calibrator merely read hedges, adding or removing words like ``maybe'' should swing its score. Instead, hedges such as ``maybe'' or ``I think'' that make a human reader judge an answer as less confident change the calibrator's score only slightly: stripping hedge words from every response shifts AUROC by $+0.0015$ and injecting them shifts it by $-0.0185$ (Table~\ref{tab:hedging}). The calibrator responds to a property of the response that the surface hedges do not control.
\paragraph{Sampling baselines without a proxy model disadvantage.} To remove the disadvantage that sampling baselines suffer from by using a proxy model, we also try drawing samples directly from LLaMA-3.1-8B and evaluate semantic entropy, self-consistency, the estimators of \citet{lin2023generating}, SPUQ~\citep{gao2024spuq}, and the conformal estimator of \citet{su2024api}. Even then, these estimators stay near chance on the hard suite, far below \textsc{Pinocchio} scored on the same responses, and drawing more samples does not narrow the gap. The same pipeline behaves as expected on easy short-form benchmarks, so the failure is specific to hard reasoning.  Detailed results are in Appendix~\ref{app:faithful_baselines}.
\subsection{Performance across benchmarks and modalities}
\label{sec:per_benchmark}\textsc{Pinocchio} performs comparably on text and vision-language responses, so modality-specific calibrators are unnecessary in this evaluation. Its accuracy varies more by task than by modality: structured mathematical benchmarks produce the clearest separation, while specialized and expert-level benchmarks are harder (Appendix Table~\ref{tab:per_benchmark_v3}).  Beyond coverage, vision-language data acts as \emph{training-signal augmentation}. Adding the vision-language set improves text AUROC most when hard text data is scarce (Appendix Table~\ref{tab:vlm_augment}); the benefit tapers as more text examples are supplied.
\begin{takeaway}A single calibrator covers both text and vision at comparable performance, and vision-language data improves calibration on text when hard text examples are scarce.
\end{takeaway}

\subsection{Cross-model transfer}
\label{sec:cross_model} We evaluate zero-shot transfer using the same calibrator. We use it to score thirteen target models absent from training, spanning eight organizations (Appendix Table~\ref{tab:cross_family}); several were released after the calibrator was trained, including GPT-6 Astra. Mean AUROC is 0.814 across the thirteen targets, a modest drop from the 0.862 it achieves on models whose responses we used during training. On the open transfer targets, where sampling baselines can be run directly, \textsc{Pinocchio} far outperforms them (Appendix~\ref{app:faithful_baselines}).
Generalization also extends to unseen benchmarks. Across four folds that each exclude five benchmarks from training, AUROC averages 0.875 on excluded benchmarks and 0.877 on included benchmarks (Appendix Table~\ref{tab:leave_k_out}).
\begin{table}[t]\centering
\footnotesize
\caption{\textbf{Lower-capability training sources improve broad transferability.} Mean AUROC is measured over eleven unseen models shared by all three ablations.}
\label{tab:capability_diversity}
\begin{tabular}{@{}lc@{}}\toprule
\vfiveprose{Training sources} & \vfiveprose{Mean transfer AUROC $\uparrow$} \\
\midrule
\vfiveprose{4 high-capability models} & \vfiveprose{0.649} \\
\vfiveprose{4 high-capability + GPT-5-mini} & \vfiveprose{0.778} \\
\vfiveprose{4 high-capability + GPT-5-mini + GPT-5.2 + Qwen3.5} & \vfiveprose{0.814} \\
\bottomrule
\end{tabular}
\end{table}
Transfer, especially to lower-capability target models improves as lower-capability source models are added. This effect is especially pronounced on individual targets such as Mistral-7B, where each successive training mixture raises transfer (Table~\ref{tab:capability_diversity}). Older models represent a domain shift in different knowledge bases and semantic cues.
\begin{table}[t]\centering
\footnotesize
\caption{\textbf{Approximately 100 labeled samples can be used to re-calibrate after zero-shot transfer.} The four open evaluation targets are absent from \textsc{Pinocchio}'s training set. We fit Platt or isotonic recalibration on 100 labeled examples and evaluate on the remainder over 25 random splits; ECE uses 15 bins.}
\label{tab:recal_unseen}
\begin{tabular}{@{}lcccc@{}}\toprule & & \multicolumn{3}{c}{ECE $\downarrow$} \\
\cmidrule(lr){3-5}Model & AUROC & Off-the-shelf & Platt & Isotonic \\
\midrule
Mistral-7B      & \vfiveprose{0.828} & \vfiveprose{0.213} & \vfiveprose{0.080} & \vfiveprose{0.059} \\
gemma-2-9b      & \vfiveprose{0.801} & \vfiveprose{0.282} & \vfiveprose{0.078} & \vfiveprose{0.055} \\
OLMo-2-7B       & \vfiveprose{0.825} & \vfiveprose{0.296} & \vfiveprose{0.066} & \vfiveprose{0.052} \\
granite-4.1-30b & \vfiveprose{0.793} & \vfiveprose{0.251} & \vfiveprose{0.081} & \vfiveprose{0.062} \\
\midrule
\textbf{Pooled} & \vfiveprose{\textbf{0.810}} & \vfiveprose{\textbf{0.259}} & \vfiveprose{\textbf{0.076}} & \vfiveprose{\textbf{0.058}} \\
\bottomrule
\end{tabular}

\end{table}

\begin{takeaway}\textsc{Pinocchio} transfers to thirteen unseen models across eight organizations.
\end{takeaway}

\subsection{Scaling and input ablations}
\label{sec:ablations}The three factors that could drive the calibrator's accuracy do not matter equally, so we measure each in turn: the amount of training data, the size of the calibrator, and the structured input it is given. Increasing the training set produces far larger gains than increasing calibrator size (Figure~\ref{fig:training_size}); calibrator size has little effect (Appendix Table~\ref{tab:model_size}), while performance keeps improving as data grows. The structured input also contributes: removing the benchmark and model-identity metadata drops AUROC from 0.863 to 0.813, so the calibrator uses more than the bare response. Full input and elicitation ablations are in Appendices~\ref{app:scaling} and \ref{app:elicitation}.  Beyond aggregate AUROC, \textsc{Pinocchio} drives three deployment workflows (adaptive clarification, confidence-gated actions, and human-escalation review), outperforming every baseline on each; Appendices~\ref{app:use_cases} and \ref{app:recalibration} detail these production use cases and per-model recalibration.
\begin{takeaway}Training data matters more than calibrator size, so uncertainty estimation performs well on a small model.
\end{takeaway}

%% file: sections/discussion.tex
\section{Using \textsc{Pinocchio}}
\label{sec:discussion}
\begin{practnote}\textbf{Adapting \textsc{Pinocchio} to a new API or distribution.} Begin with the released checkpoint and evaluate it on a labeled sample from the intended deployment. If response ranking transfers but the probabilities are shifted, fit Platt or isotonic regression on a labeled calibration set. If ranking performance also degrades, fine-tuning is appropriate.
\end{practnote}

\paragraph{Incorporating \textsc{Pinocchio} in your code.}The released \texttt{pinocchio-uq} package scores the output of an existing API call without reading the target model's logits or weights. After installing it with \texttt{pip install pinocchio-uq}, the minimal integration is:{\small
\begin{verbatim}
from pinocchio import Pinocchio
judge = Pinocchio()  # load once

response = client.chat.completions.create(
    model="gpt-5", messages=messages
)
p_correct = judge.score(response, messages=messages)
\end{verbatim}
}
\paragraph{Limitations.} The calibrator requires labeled training data with known ground-truth correctness, limiting applicability to tasks where automated grading is possible. Cross-family transfer degrades compared to within-family (Section~\ref{sec:cross_model}). Including benchmark and model-identity metadata raises AUROC from 0.813 to 0.863 which may limit performance in settings where that metadata is unknown. Our training data is English-only. The calibrator is weakest on the most difficult benchmarks where even correct responses hedge and qualify, the same markers associated with incorrectness (Appendix~\ref{app:per_benchmark}).
\section*{Acknowledgements} This project was supported by the Center for AI and Responsible Financial Innovation at Columbia University through a research award, and by the NVIDIA Academic Grant Program.

%% file: sections/appendix.tex
\section{Training data details}
\label{app:data}\vfiveprose{Table~\ref{tab:data_unified} shows the training data breakdown for \textsc{Pinocchio}. Seven models generated 32{,}419 training responses across 20 benchmarks: 17{,}778 vision-language examples and 14{,}641 text examples.}
\begin{table}[t]\centering
\caption{\vfiveprose{\textbf{The training set contains responses from seven models.} Exact training counts by model.}}
\label{tab:data_unified}\small
\begin{tabular}{@{}lr@{}}\toprule
\textbf{Model} & \textbf{Training examples} \\
\midrule
GPT-5-mini & \vfiveprose{4{,}584} \\
GPT-5.2 & \vfiveprose{4{,}618} \\
Qwen3.5-397B & \vfiveprose{3{,}712} \\
GPT-5.6 & \vfiveprose{4{,}960} \\
Claude Fable~5 & \vfiveprose{4{,}892} \\
Kimi~3 & \vfiveprose{4{,}697} \\
Claude Opus~5 & \vfiveprose{4{,}956} \\
\midrule
\textbf{Total} & \vfiveprose{\textbf{32{,}419}} \\
\bottomrule
\end{tabular}

\end{table}

\section{Hyperparameters}
\label{app:hyper}Table~\ref{tab:hyper} lists training hyperparameters for \textsc{Pinocchio}.
\begin{table}[t]\centering
\caption{\textbf{The calibrator is trained with lightweight LoRA fine-tuning.} \textsc{Pinocchio} training hyperparameters.}
\label{tab:hyper}
\begin{tabular}{@{}ll@{}}\toprule
Hyperparameter & Value \\
\midrule
Base model & Qwen3-VL-8B-Instruct \\
LoRA rank ($r$) & 32 \\
LoRA alpha ($\alpha$) & 64 \\
LoRA target modules & All linear layers \\
Learning rate & $1 \times 10^{-4}$ \\
Weight decay & 0.01 \\
Batch size & 1 \\
Gradient accumulation steps & 16 \\
Effective batch size & 16 \\
Epochs & 3 \\
Max question length & 1{,}500 characters \\
Max response length & 800 characters \\
Train/test split & 85/15 (question-level split) \\
Precision & bf16 \\
Optimizer & AdamW \\
Hardware & 4$\times$ GPU (DDP) \\
Training time & 2--3 hours \\
\bottomrule
\end{tabular}

\end{table}

\section{Training configuration and input processing}
\label{app:training_config}
\paragraph{Training configuration.}Training uses AdamW with learning rate $10^{-4}$, batch size 1 with gradient accumulation over 16 steps (effective batch size 16), BF16 mixed precision, and runs for 3 epochs on 4 GPUs with DDP. The full training takes approximately 2--3 hours on A100 GPUs. See Table~\ref{tab:hyper} for complete hyperparameters.
\paragraph{Truncation and image processing.}We truncate questions to 1{,}500 characters and responses to 800 characters, and report truncation results in Appendix~\ref{app:scaling}. For VLM benchmarks, we resize real images to fit within 200{,}704 to 401{,}408 pixels (corresponding to 256--512 tiles of $28 \times 28$ in the model's dynamic-resolution tiling scheme). For text-only benchmarks, we use a $28 \times 28$ gray placeholder image.
\section{Training data details: multi-model training, benchmark selection, and data sufficiency}
\label{app:training_data_details}
\paragraph{Why multi-model training?}\newprose{An earlier source-diversity ablation uses more responses from a broader set of sources than any single-source run. Single-source calibrators obtain 0.672--0.759 AUROC on held-out target models, while the pooled calibrator reaches 0.878 (Section~\ref{sec:cross_model}). This comparison changes both sample count and source diversity, so it does not isolate a mechanism for the improvement. In leave-one-model-out evaluation, training on two sources and testing on the third yields mean AUROC 0.843, 0.035 below the full model. This directly measures transfer to an excluded source model.}
\paragraph{Training data sufficiency and split integrity.}The training size ablation in Appendix~\ref{app:scaling} shows that 5{,}000 samples achieve 96\% of full-data performance and 2{,}000 samples achieve 91\%. We enforce a \emph{question-level split}: \newprose{when multiple models answer the same question in the matched ablation set, we assign all instances to the same split.} This is essential: a na\"ive sample-level split \orangeprose{would place near-identical instances of the same question in both train and test}. All results in this paper use the question-level split with zero question overlap between train and test.
\section{Additional analyses}
\label{app:additional}
\subsection{Full results tables}\newprose{The matched baseline comparison is reported in the main text (Table~\ref{tab:main}, Section~\ref{sec:main_results}), where \textsc{Pinocchio} reaches \purpleprose{0.863} AUROC. \purpleprose{Table~\ref{tab:frontier_flagship} gives its per-model held-out evaluation on the models whose responses trained it}, and Table~\ref{tab:cross_family} reports its zero-shot transfer to \orangeprose{thirteen} unseen models.}
\begin{table}[t]\centering
\footnotesize
\caption{\newprose{\textbf{A single calibrator discriminates and calibrates held-out \purpleprose{responses from the models whose responses trained it}.}} \purpleprose{Length AUROC is the response-length baseline (in-sample $\log(1+\text{length})$ logistic fit) on the same held-out responses; the Overall value is the sample-weighted mean of the per-model baselines. \textsc{Pinocchio} exceeds it on every model.}}
\label{tab:frontier_flagship}
\begin{tabular}{@{}lcccc@{}}\toprule
Target model & AUROC $\uparrow$ & \purpleprose{Length AUROC $\uparrow$} & \newprose{Brier $\downarrow$} & \newprose{ECE $\downarrow$} \\
\midrule
Claude Fable~5 & \vfiveprose{0.852} & \purpleprose{0.575} & \vfiveprose{0.165} & \vfiveprose{0.135} \\
GPT-5.6 & \vfiveprose{0.852} & \purpleprose{0.598} & \vfiveprose{0.174} & \vfiveprose{0.141} \\
Claude Opus~5 & \vfiveprose{0.859} & \purpleprose{0.607} & \vfiveprose{0.163} & \vfiveprose{0.129} \\
Kimi~3 & \vfiveprose{0.862} & \purpleprose{0.600} & \vfiveprose{0.168} & \vfiveprose{0.134} \\
\midrule
\textbf{Overall} & \vfiveprose{\textbf{0.856}} & \purpleprose{0.595} & \vfiveprose{0.167} & \vfiveprose{\textbf{0.131}} \\
\bottomrule
\end{tabular}
\end{table}

\begin{table}[t]\centering
\footnotesize
\caption{\vfiveprose{\textbf{\textsc{Pinocchio} transfers to \orangeprose{thirteen} unseen models.} Zero-shot transfer AUROC on target models absent from training, spanning \orangeprose{eight} organizations. Models marked $^\dagger$ were released after training.} \purpleprose{Length AUROC is the response-length baseline (in-sample $\log(1+\text{length})$ logistic fit) on the same responses; \textsc{Pinocchio} exceeds it on every target. LLaMA-3.1-8B is omitted from this column (its responses are evaluated on a separate hard-suite split).} \orangeprose{GPT-6 Astra is scored zero-shot on its held-out split ($n{=}380$); the other targets use the full transfer bundle.}}
\label{tab:cross_family}
\begin{tabular}{@{}llcc@{}}\toprule
Model & Organization & AUROC $\uparrow$ & \purpleprose{Length AUROC $\uparrow$} \\
\midrule
\orangeprose{GPT-6 Astra}$^{\dagger}$ & \orangeprose{OpenAI} & \orangeprose{0.804} & \orangeprose{0.633} \\
LLaMA-3.1-8B                       & Meta      & \vfiveprose{0.825} & \purpleprose{---} \\
Mistral-7B                         & Mistral   & \vfiveprose{0.828} & \purpleprose{0.642} \\
Devstral-2-24B$^\dagger$           & Mistral   & \vfiveprose{0.826} & \purpleprose{0.608} \\
gemma-2-9b                         & Google    & \vfiveprose{0.801} & \purpleprose{0.645} \\
OLMo-2-7B                          & AllenAI   & \vfiveprose{0.825} & \purpleprose{0.585} \\
Olmo-3.1-32B$^\dagger$             & AllenAI   & \vfiveprose{0.793} & \purpleprose{0.623} \\
DeepSeek-R1-Distill-32B            & DeepSeek  & \vfiveprose{0.802} & \purpleprose{0.515} \\
granite-4.1-30b$^\dagger$          & IBM       & \vfiveprose{0.793} & \purpleprose{0.621} \\
Claude Opus~4.6                    & Anthropic & \vfiveprose{0.783} & \purpleprose{0.671} \\
Claude Sonnet~4.6                  & Anthropic & \vfiveprose{0.807} & \purpleprose{0.669} \\
Gemini 3.1 Pro                     & Google    & \vfiveprose{0.863} & \purpleprose{0.580} \\
Gemini 3 Flash                     & Google    & \vfiveprose{0.832} & \purpleprose{0.583} \\
\bottomrule
\end{tabular}

\end{table}

\subsection{Cross-model training}Table~\ref{tab:cross_model} shows that training on multiple source models, rather than a single one, is what makes the calibrator transfer: each single-source calibrator is evaluated on held-out target models it never trained on.
\begin{table}[t]\centering
\footnotesize
\caption{\textbf{Training on multiple source models is what makes the calibrator transfer.} Cross-model transfer. AUROC when training on single vs.\ multiple source models. \vfiveprose{Multi-model training yields an AUROC gain of 0.12--0.21 over single-source training.} Single-source rows evaluate on held-out target models the calibrator never trained on (the diagonal is N/A).}
\label{tab:cross_model}
\begin{tabular}{@{}lcccc@{}}\toprule
Training data & GPT-5-mini & GPT-5.2 & Qwen3.5 & Mean \\
\midrule
GPT-5-mini only & N/A & 0.715 & 0.693 & 0.704 \\
GPT-5.2 only & 0.741 & N/A & 0.776 & 0.759 \\
Qwen3.5 only & 0.658 & 0.687 & N/A & 0.672 \\
\midrule
\purpleprose{All three sources} & \textbf{0.882} & \textbf{0.877} & \textbf{0.873} & \textbf{0.878} \\
\bottomrule
\end{tabular}

\end{table}

\subsection{Input and robustness ablations}Table~\ref{tab:knowledge_cues} decomposes what the calibrator reads (the question versus the response), and Table~\ref{tab:hedging} shows that stripping or injecting explicit hedge words barely moves its score.
\begin{table}[t]\centering
\footnotesize
\caption{\textbf{The calibrator draws on both the question and the answer.} \purpleprose{Blanking or redacting the question leaves most of the signal intact, so the calibrator reads the answer rather than only gauging question difficulty, while the remaining drop shows the question still contributes.} Brackets give 95\% bootstrap CIs. ``Empty question'' leaves the question field blank; ``question redacted'' replaces it with the literal string ``[QUESTION REDACTED]''; \purpleprose{both use the trained calibrator without retraining, changing only the inference-time question field.}}
\label{tab:knowledge_cues}
\begin{tabular}{@{}lcc@{}}\toprule
Input & AUROC $\uparrow$ & $\Delta$ \\
\midrule
Question + response (canonical)      & \purpleprose{0.863} & -- \\
Response only, empty question        & \purpleprose{0.805} & \purpleprose{$-0.058$} \\
Response only, question redacted     & \purpleprose{0.795} & \purpleprose{$-0.068$} \\
\bottomrule
\end{tabular}

\end{table}

\begin{table}[t]\centering
\footnotesize
\caption{\textbf{Explicit hedge words barely move the calibrator's score.} Hedging probe. Stripping or injecting explicit hedge words (``maybe'', ``I think'', ``possibly'') barely changes AUROC; the calibrator uses signals beyond superficial hedging. The last column is the mean shift in predicted $\pcorrect$.}
\label{tab:hedging}
\begin{tabular}{@{}lccc@{}}\toprule
Condition & AUROC & $\Delta$ vs original & Mean shift in $\pcorrect$ \\
\midrule
Original responses & \purpleprose{0.8633} & -- & -- \\
Hedging stripped   & \purpleprose{0.8642} & \purpleprose{$+0.0009$} & \purpleprose{$+0.0002$} \\
Hedging injected   & \purpleprose{0.8573} & \purpleprose{$-0.0060$} & \purpleprose{$-0.056$} \\
\bottomrule
\end{tabular}

\end{table}

\subsection{Per-benchmark breakdown}
\label{app:per_benchmark}Table~\ref{tab:per_benchmark_v3} reports per-benchmark AUROC and effect size on the held-out test set.
\begin{table}[t]\centering
\caption{\textbf{The calibrator discriminates correctness across every benchmark.} Per-benchmark calibrator performance on the held-out test set (question-level split, zero question overlap with training). $d$ = Cohen's $d$ between calibrator scores for correct vs.\ incorrect responses. Sorted by AUROC descending.}
\label{tab:per_benchmark_v3}\small
\begin{tabular}{llrccr}\toprule
\textbf{Benchmark} & \textbf{Modality} & \textbf{$N$} & \textbf{Acc} & \textbf{AUROC} & \textbf{$d$} \\
\midrule
LiveBench       & Text & \purpleprose{91}  & \purpleprose{.637} & \purpleprose{.958} & \purpleprose{2.58} \\
MathVista       & VLM  & \purpleprose{116} & \purpleprose{.397} & \purpleprose{.944} & \purpleprose{2.55} \\
ARC-AGI         & Text & \purpleprose{114} & \purpleprose{.421} & \purpleprose{.935} & \purpleprose{2.54} \\
RealWorldQA     & VLM  & \purpleprose{112} & \purpleprose{.491} & \purpleprose{.931} & \purpleprose{1.99} \\
BBEH            & Text & \purpleprose{125} & \purpleprose{.456} & \purpleprose{.877} & \purpleprose{1.58} \\
OmniMath        & Text & \purpleprose{78}  & \purpleprose{.282} & \purpleprose{.871} & \purpleprose{1.73} \\
MM-Vet          & VLM  & \purpleprose{73}  & \purpleprose{.342} & \purpleprose{.863} & \purpleprose{1.67} \\
HallusionBench  & VLM  & \purpleprose{111} & \purpleprose{.847} & \purpleprose{.861} & \purpleprose{1.78} \\
MathVerse       & VLM  & \purpleprose{103} & \purpleprose{.505} & \purpleprose{.857} & \purpleprose{1.38} \\
MathVision      & VLM  & \purpleprose{88}  & \purpleprose{.273} & \purpleprose{.799} & \purpleprose{1.09} \\
PRBench         & Text & \purpleprose{78}  & \purpleprose{.769} & \purpleprose{.775} & \purpleprose{1.15} \\
SimpleQA        & Text & \purpleprose{120} & \purpleprose{.208} & \purpleprose{.773} & \purpleprose{1.09} \\
VizWiz          & VLM  & \purpleprose{96}  & \purpleprose{.656} & \purpleprose{.769} & \purpleprose{1.07} \\
HLE             & Text & \purpleprose{94}  & \purpleprose{.170} & \purpleprose{.766} & \purpleprose{1.20} \\
MMStar          & VLM  & \purpleprose{116} & \purpleprose{.741} & \purpleprose{.727} & \purpleprose{0.64} \\
ChemBench       & Text & \purpleprose{108} & \purpleprose{.731} & \purpleprose{.671} & \purpleprose{0.69} \\
GPQA            & Text & \purpleprose{90}  & \purpleprose{.711} & \purpleprose{.653} & \purpleprose{0.48} \\
CharXiv         & VLM  & \purpleprose{118} & \purpleprose{.678} & \purpleprose{.651} & \purpleprose{0.45} \\
MMMU            & VLM  & \purpleprose{66}  & \purpleprose{.879} & \purpleprose{.630} & \purpleprose{0.60} \\
HLE-Multimodal  & VLM  & \purpleprose{58}  & \purpleprose{.190} & \purpleprose{.557} & \purpleprose{0.02} \\
\midrule
\textbf{Mean}   & --  & \textbf{\purpleprose{98}} & \textbf{\purpleprose{.523}} & \textbf{\purpleprose{.793}} & \textbf{\purpleprose{1.31}} \\
\bottomrule
\end{tabular}

\end{table}

\subsection{Response length analysis}
\begin{table}[t]\centering
\caption{\textbf{Response length explains little of the calibrator's signal.} \purpleprose{Response length analysis on the matched set. A length-only predictor reaches far below the calibrator, so response length alone is only weakly predictive of correctness.}}
\label{tab:length_analysis_v3}\small
\begin{tabular}{lr}\toprule
\textbf{Metric} & \textbf{Value} \\
\midrule
Calibrator AUROC & \purpleprose{0.863} \\
Length-only baseline AUROC & \purpleprose{0.570} \\
\bottomrule
\end{tabular}

\end{table}
\purpleprose{The response-length baseline in Table~\ref{tab:main} (0.566 AUROC) and the length-only entry above (0.570) both show that response length alone is only weakly predictive of correctness.}
\subsection{Difficulty stratification}Table~\ref{tab:difficulty_v3} breaks down calibrator performance by benchmark difficulty tier.
\begin{table}[t]\centering
\caption{\purpleprose{\textbf{Calibrator AUROC is lower on easy benchmarks, where errors are scarce.}} Calibrator performance by benchmark difficulty tier.}
\label{tab:difficulty_v3}\small
\begin{tabular}{lrrcr}\toprule
\textbf{Tier} & \textbf{Benchmarks} & \textbf{$N$} & \textbf{Mean acc} & \textbf{AUROC} \\
\midrule
Easy ($\geq$0.65)   & 8 &  \purpleprose{783} & \purpleprose{0.746} & \purpleprose{0.724} \\
Medium (0.40--0.65) & 5 &  \purpleprose{545} & \purpleprose{0.495} & \purpleprose{0.909} \\
Hard ($<$0.40)      & 7 &  \purpleprose{627} & \purpleprose{0.270} & \purpleprose{0.840} \\
\midrule
\textbf{All}        & \textbf{20} & \textbf{\purpleprose{1{,}955}} & \textbf{\purpleprose{0.523}} & \textbf{\purpleprose{0.863}} \\
\bottomrule
\end{tabular}

\end{table}

\subsection{Difficulty-controlled disagreement test}
\label{app:disagreement}\purpleprose{On very hard questions a calibrator can look accurate without reading the answer at all: if almost every model fails a question, always predicting ``incorrect'' is right most of the time. To rule this out, we keep only the questions where the models whose responses trained \textsc{Pinocchio} \emph{disagree}, with at least one correct and at least one incorrect response (397 questions, $n{=}2{,}389$ graded responses). Within one such question every response shares the same question, so difficulty is held fixed: the only thing separating a correct response from an incorrect one is the response itself, and any predictor that does not read the response must give every response to the question the same score and so cannot rank them. Averaged within question, \textsc{Pinocchio} ranks the correct responses above the incorrect ones with AUROC 0.726, well above the 0.5 an answer-blind predictor is pinned to.

We confirm this is not \orangeprose{a chance effect} of \textsc{Pinocchio}'s score distribution with a within-question permutation test. Holding \textsc{Pinocchio}'s scores fixed, we repeatedly reassign the correct/incorrect labels among the responses to each question, keeping the number of correct responses per question unchanged, and recompute the overall AUROC each time. Because the reshuffling stays within questions, it preserves each question's difficulty and traces out exactly the AUROC an answer-blind predictor could reach by chance. \textsc{Pinocchio}'s pooled AUROC of 0.742 is higher than under every one of these relabelings ($p<0.0001$), so its ordering of responses tracks which one is actually correct, not merely which questions are hard.

We further check that \textsc{Pinocchio} responds to the answer rather than the model-identity tag. For each disagreement question we re-score every response after swapping in the answers the other source models gave to the same question, holding the original slot's model tag fixed. Across all 14{,}687 (answer, slot) pairs,  \textsc{Pinocchio}'s score follows the swapped-in answer's correctness at AUROC 0.705 (chance 0.5): even when an answer is paired with a different model's tag, the calibrator ranks correct answers above incorrect ones, so it reads the response rather than the tag.}

\subsection{Leave-$K$-out cross-validation}Table~\ref{tab:leave_k_out} reports per-fold results for leave-$K$-out cross-validation ($K=5$, 4 folds). The mean gap between included- and excluded-benchmark AUROC is 0.002.
\begin{table}[t]\centering
\small
\caption{\textbf{Held-out benchmarks match in-distribution performance.} Leave-$K$-out cross-validation ($K=5$, 4 folds, question-level split). In-dist.\ and held-out AUROC for each fold. Gap = in-dist.\ minus held-out (positive = degradation on unseen benchmarks).}
\label{tab:leave_k_out}\resizebox{\textwidth}{!}{
\begin{tabular}{@{}clccc@{}}\toprule
Fold & Held-out benchmarks & In-dist. & Held-out & Gap \\
\midrule0 & GPQA, Hallusion., MathVerse, MM-Vet, VizWiz & 0.879 & 0.866 & +0.013 \\
1 & HLE, MMMU, MMStar, OmniMath, SimpleQA & 0.873 & 0.885 & $-$0.012 \\
2 & BBEH, CharXiv, MathVision, MathVista, RealWorldQA & 0.878 & 0.876 & +0.002 \\
3 & ARC-AGI, ChemBench, HLE-Multimodal, LiveBench, PRBench & 0.879 & 0.872 & +0.007 \\
\midrule
\multicolumn{2}{l}{Mean $\pm$ std} & 0.877 & 0.875 & +0.002 $\pm$ 0.009 \\
\bottomrule
\end{tabular}
}
\end{table}

\subsection{Elicitation strategy ablation}
\label{app:elicitation}Table~\ref{tab:elicitation} compares different ways to extract the uncertainty signal from calibrator checkpoints on an internal validation split (absolute values therefore differ from the final model in Table~\ref{tab:main}). The standard logit-based approach (softmax over the \texttt{i}/\texttt{ii} tokens) achieves 0.859 AUROC. MC Dropout~\citep{gal2016dropout} (N=5 forward passes) provides no improvement (0.859), consistent with LoRA's low-rank perturbations providing insufficient stochasticity. Verbalized probability (prompting the calibrator to output a number) degrades to 0.749. A hidden-state MLP probe achieves 0.878 (\vfiveprose{+0.019 AUROC}), but requires open-weight access to the calibrator's internals. Temperature scaling ($T = 1.34$) leaves AUROC unchanged while marginally improving ECE (0.023 to 0.019), indicating the model is already well-calibrated.
\begin{table}[t]\centering
\small
\caption{\textbf{Simple logit-based elicitation captures nearly all available signal.} Elicitation strategy ablation evaluated on an internal validation split. Absolute AUROC values differ from Table~\ref{tab:main} (which reports the final model on the held-out test set); relative comparisons between strategies are unchanged. Logit-based extraction captures 95\%+ of the available signal.}
\label{tab:elicitation}
\begin{tabular}{lc}\toprule
Strategy & AUROC \\
\midrule
Logit (softmax over i/ii) & 0.859 \\
MC Dropout (N=5) & 0.859 \\
Hidden-state MLP probe & 0.878 \\
Hidden + logit combined & 0.884 \\
Verbalized probability & 0.749 \\
Token entropy & 0.520 \\
\bottomrule
\end{tabular}

\end{table}

\subsection{Off-the-shelf decision models}
\label{app:decision_models}
\redprose{A natural question is whether a purpose-built \emph{decision model}, which returns calibrated probabilities over a fixed answer space in a single forward pass, can predict black-box correctness without task-specific training. We evaluate TypeSafe Jev~\citep{typesafe2026jev}, a commercial API of this kind, queried through its native typed-question interface with the question and response as state, under the same truncation as \textsc{Pinocchio}.}

\redprose{Jev does not match a trained calibrator. Scored on the same held-out responses and labels, it reaches 0.666 AUROC (95\% CI [0.638, 0.691]) with Brier 0.240 and ECE 0.133 over 1{,}718 text responses, against 0.868 for \textsc{Pinocchio} on the identical rows. The gap holds on every benchmark we test, ranging from 0.046 AUROC on GPQA to 0.375 on LiveBench, and Jev is strongest where answers are short and factual (SimpleQA 0.687, PRBench 0.685) and weakest on multi-step reasoning (LiveBench 0.574, HLE 0.594). Jev does not accept images, so the evaluation covers the text-only subset. It is also non-deterministic, returning different scores for identical repeated requests, which adds roughly 0.02 of run-to-run variation to any single-draw AUROC.}

\section{Per-model results}
\label{app:per_model}Table~\ref{tab:per_model_main} shows per-model AUROC for the main comparison.
\begin{table}[t]\centering
\caption{\textbf{\textsc{Pinocchio} outperforms baselines consistently across target models.} Per-model AUROC on held-out test set. \textsc{Pinocchio} is consistent across all target models.}
\label{tab:per_model_main}\small
\begin{tabular}{@{}lccc@{}}\toprule
Method & GPT-5-mini & GPT-5.2 & Qwen3.5 \\
\midrule
Verbalized (raw) & 0.645 & 0.637 & 0.569 \\
Verbalized (Isotonic) & 0.647 & 0.646 & 0.659 \\
Response length & 0.633 & 0.609 & 0.500 \\
Combined & 0.693 & 0.658 & 0.656 \\
\midrule
\textsc{Pinocchio} & \purpleprose{\textbf{0.886}} & \purpleprose{\textbf{0.858}} & \purpleprose{\textbf{0.835}} \\
\bottomrule
\end{tabular}

\end{table}

\section{Production use case details}
\label{app:use_cases}
\label{sec:use_cases}
\begin{figure}[h!]  \centering
\includegraphics[width=0.65\textwidth]{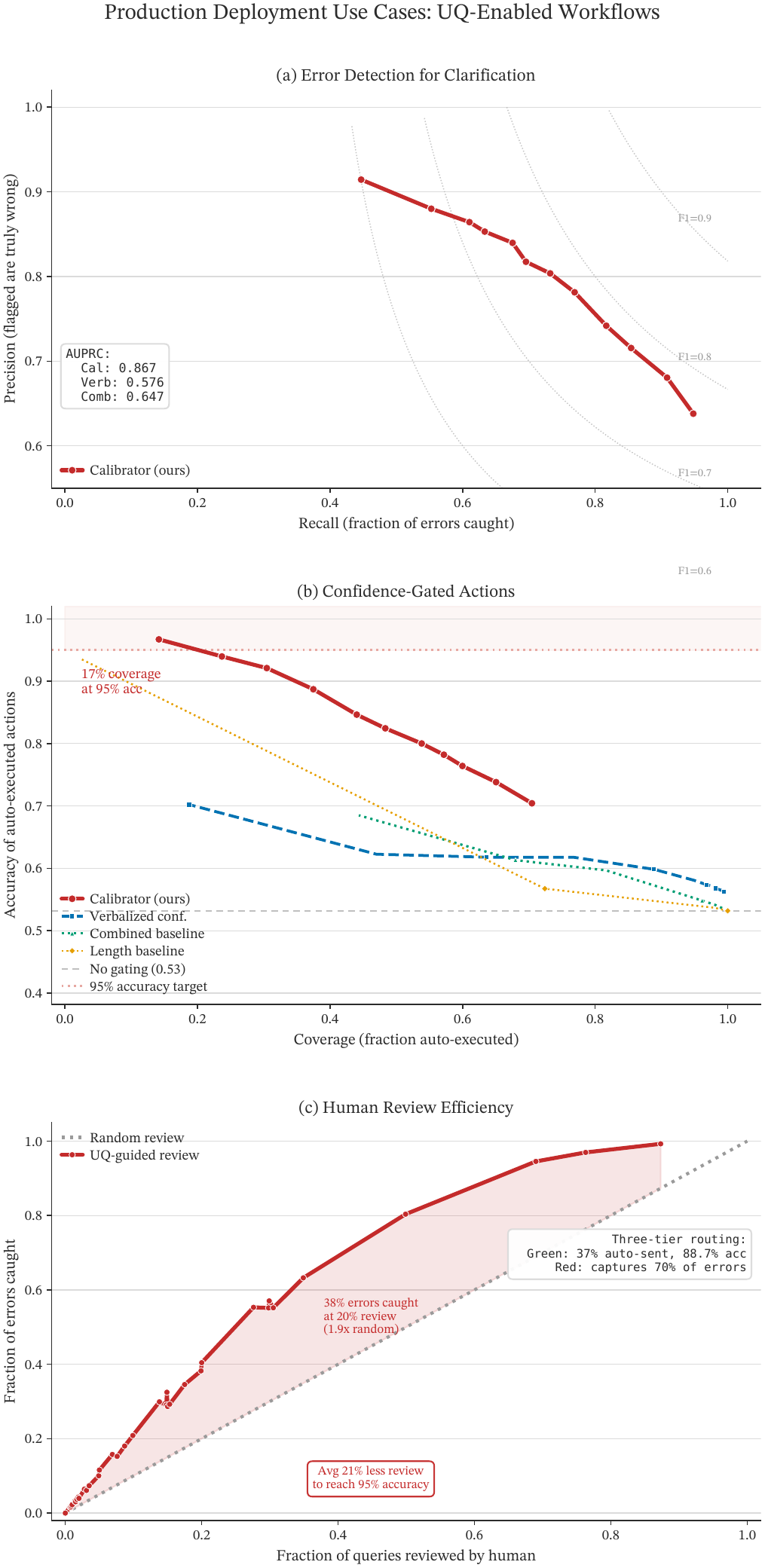}
\caption{\textbf{Across three production workflows, the calibrator beats every baseline.} (a)~Error detection: calibrator AUPRC = 0.867 vs.\ verbalized 0.576. (b)~Confidence-gated actions: only the calibrator achieves meaningful coverage at $\geq$90\% accuracy. (c)~Human escalation: UQ-guided review reduces workload by 21\% to reach 95\% accuracy.}
\label{fig:production}
\end{figure}

AUROC measures discrimination in aggregate, but practitioners need to know: \emph{what can I actually do with a calibrator score?} We evaluate three deployment scenarios on the same held-out test set (1{,}953 examples), comparing the calibrator against every baseline from the main results.
\paragraph{Adaptive clarification (error detection).}Consider a customer-facing chatbot where incorrect responses damage user trust. The system monitors each response and flags likely errors for human review before delivery. We measure this as a binary detection task: given a response, predict whether it is incorrect. The calibrator achieves AUPRC of 0.867 and best F1 of 0.782 across three target models, compared to 0.576 AUPRC for verbalized confidence (Table~\ref{tab:uc_e}). We find numerous examples where the target model reports 100\% verbalized confidence while the calibrator correctly assigns $\pcorrect < 0.01$; all such cases are indeed incorrect (Figure~\ref{fig:production}a).
\begin{table}[h]  \centering  \small
\caption{\textbf{\textsc{Pinocchio} detects incorrect responses far better than baselines.} Error detection for adaptive clarification. AUPRC and best F1 for detecting incorrect responses, averaged across three target models.}
\label{tab:uc_e}
\begin{tabular}{lcc}    \toprule    Method & AUPRC & Best F1 \\
    \midrule    \textsc{Pinocchio} (ours) & \textbf{0.867} & \textbf{0.782} \\
    Verbalized confidence & 0.576 & 0.405 \\
    Combined baseline & 0.647 & 0.650 \\
    Isotonic (verbalized) & 0.628 & 0.637 \\
    Length baseline & 0.534 & 0.642 \\
    \bottomrule
\end{tabular}

\end{table}

\paragraph{Confidence-gated actions (agentic safety).}Consider an agentic pipeline where an LLM takes irreversible actions (e.g., sending emails, executing trades, modifying databases). Only responses above a confidence threshold $\pcorrect > t$ are auto-executed; the rest are held for human verification. The calibrator achieves 27--35\% auto-execution coverage at 90\% accuracy across target models; no baseline achieves comparable coverage at the same accuracy (Figure~\ref{fig:production}b).
\paragraph{Human-in-the-loop escalation (enterprise triage).}Consider an enterprise helpdesk where thousands of queries arrive daily but only a limited number of human reviewers are available. By reviewing low-confidence responses first (ranked by calibrator score), reviewers catch errors faster than random ordering. To reach 95\% system accuracy, UQ-guided review requires examining only 79\% of responses vs.\ 100\% with random ordering, a 21\% workload reduction (Figure~\ref{fig:production}c). A three-tier routing (high/medium/low confidence) assigns 35--43\% of queries to auto-delivery at 88--90\% accuracy, containing only 9\% of total errors (Table~\ref{tab:uc_g}).

\subsection{Per-model use case breakdown}
\begin{figure}[t]  \centering
\includegraphics[width=0.85\textwidth]{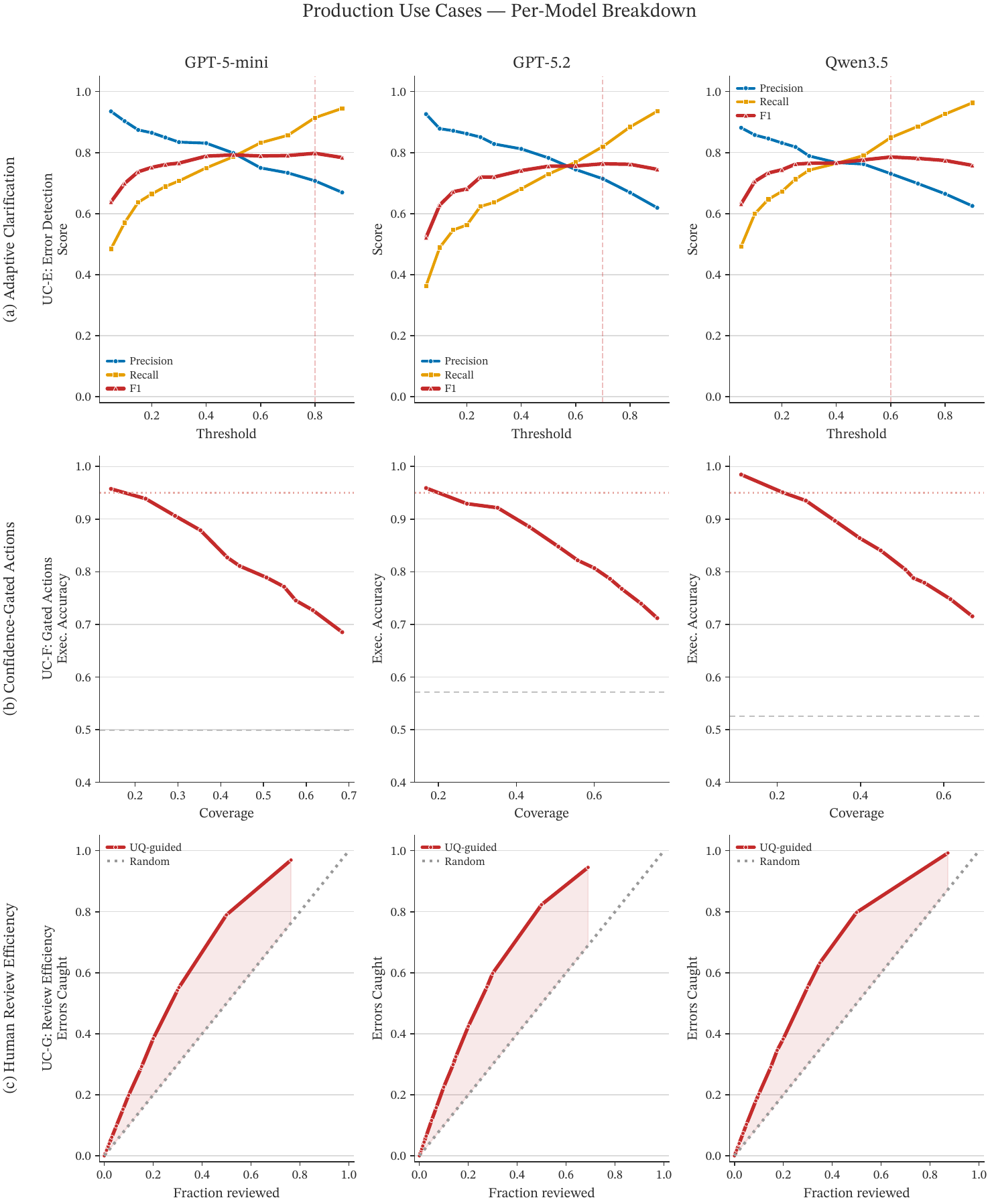}
\caption{\textbf{The three production use cases hold consistently across every target model.} Top row: error detection (precision, recall, F1 vs.\ threshold). Middle row: coverage vs.\ accuracy for confidence-gated actions. Bottom row: human review efficiency (UQ-guided vs.\ random). Columns: GPT-5-mini, GPT-5.2, Qwen3.5. Results are consistent across all target models.}
\label{fig:production_permodel}
\end{figure}

Figure~\ref{fig:production_permodel} and Tables~\ref{tab:uc_f}--\ref{tab:uc_g} break down all three use cases by target model.

\begin{table}[t]  \centering  \small
\caption{\textbf{The calibrator enables meaningful auto-execution coverage at strict error targets.} Confidence-gated actions: coverage at target accuracy levels, per model (\vfiveprose{held-out test set}, $n = 1{,}953$).}
\label{tab:uc_f}
\begin{tabular}{lccc}    \toprule    Risk Profile & GPT-5-mini & GPT-5.2 & Qwen3.5 \\
    \midrule    Aggressive ($\leq$10\% error) & 29.3\% & 35.2\% & 26.9\% \\
    Moderate ($\leq$5\% error) & 14.4\% & 16.8\% & 21.3\% \\
    \bottomrule
\end{tabular}

\end{table}

\begin{table}[t]  \centering  \small
\caption{\textbf{Confidence-based routing cuts review workload while concentrating errors in the red tier.} Human escalation: three-tier routing and workload reduction, per model (question-level split).}
\label{tab:uc_g}
\begin{tabular}{lccc}    \toprule    & GPT-5-mini & GPT-5.2 & Qwen3.5 \\
    \midrule    Green tier volume & 35\% & 43\% & 34\% \\
    Green tier accuracy & 0.879 & 0.885 & 0.897 \\
    Green tier error share & 9\% & 12\% & 7\% \\
    Red tier error share & 71\% & 64\% & 74\% \\
    Review for 95\% acc (UQ) & 78\% & 79\% & 79\% \\
    Review for 95\% acc (random) & 100\% & 100\% & 100\% \\
    Workload reduction & 22\% & 21\% & 21\% \\
    \bottomrule
\end{tabular}

\end{table}

\subsection{Additional downstream applications}
\label{app:additional_use_cases}Beyond the three primary use cases above, we briefly evaluate \textsc{Pinocchio} on additional downstream tasks:
\paragraph{DPO pair selection.}In reinforcement learning from human feedback (RLHF), training requires pairs of preferred and dispreferred responses. We use the calibrator's confidence gap between two responses to the same question to select \emph{informative} pairs where the calibrator is confident one is correct and the other incorrect. This achieves 88.8\% informative pair accuracy, compared to random pairing which yields many uninformative pairs where both responses are correct or both incorrect.
\paragraph{Model selection.}Given $N$ candidate models, we use the calibrator to select the most confident model's response for each query. With $N{=}3$ models, this achieves 67.4\% accuracy, a +5.4\% improvement over always using the single best model. The calibrator effectively routes each query to the model most likely to answer it correctly.
\paragraph{Data filtering.}We use calibrator confidence to filter training data, retaining only high-confidence samples for downstream fine-tuning. At 50\% retention, filtered accuracy reaches 90.9\%, \purpleprose{much} higher than the unfiltered baseline. This suggests \textsc{Pinocchio} can serve as a quality filter for synthetic data pipelines.
\paragraph{Step-level uncertainty.}We attempted to apply the calibrator at the reasoning-step level (predicting whether individual chain-of-thought steps are correct). This achieved only 0.53 AUROC, near random, indicating that the calibrator's signal is response-level rather than step-level. Developing step-level uncertainty estimation remains an open problem.
\section{Domain-specific deployment details}
\label{app:domain_deployment}We evaluate \textsc{Pinocchio} in \vfiveprose{two} domain-specific scenarios to illustrate how calibrated uncertainty enables practical deployment decisions. In each case, we simulate a tiered routing system where the calibrator's confidence score determines whether an LLM response is auto-delivered, reviewed by a junior expert, or escalated to a senior expert.
\subsection{Healthcare deployment}Consider a telehealth platform where patients submit medical questions to an LLM. Incorrect medical advice can cause direct harm, so confidence-based routing is critical: high-confidence responses can be sent directly, moderate-confidence responses reviewed by a nurse, and low-confidence responses escalated to a physician. We evaluate on 1{,}006 samples from medical/scientific benchmarks (ChemBench, MMMU, HLE, GPQA, SimpleQA), achieving 0.898 AUROC with a base accuracy of 54.9\%.
\paragraph{Clinical triage.}\newprose{Table~\ref{tab:health_triage} is an illustrative threshold analysis, not a clinical study. At threshold 0.9, 4.1\% of auto-approved benchmark responses are wrong and coverage is 27\%. At threshold 0.7, coverage is 43\%. The 90\% ``resident'' row is an assumed comparison point rather than a measured human baseline and must not be interpreted as clinical evidence.}
\begin{table}[t]\centering
\small
\caption{\newprose{\textbf{Confidence thresholds trade coverage for error rate on the healthcare benchmark subset.} This is an illustrative threshold analysis, not a clinical evaluation; no human baseline was measured.}}
\label{tab:health_triage}
\begin{tabular}{@{}cccc@{}}\toprule
Threshold & Coverage & Auto Error Rate & Catch Rate \\
\midrule0.5 & 59\% & 19.3\% & 75\% \\
0.6 & 50\% & 13.2\% & 85\% \\
0.7 & 43\% & 10.4\% & 90\% \\
0.8 & 34\% & 7.7\% & 94\% \\
0.9 & 27\% & 4.1\% & 98\% \\
\bottomrule
\end{tabular}

\end{table}

\paragraph{Harm severity.}Of 454 total errors, only 26 (5.7\%) fall in the high-confidence tier ($p > 0.8$), while 340 (74.9\%) are correctly assigned low confidence ($p < 0.5$) and would be flagged automatically. Reviewing the least-confident responses yields a number-needed-to-review (NNR) of $\approx$1.0 for the first 25 reviews, meaning nearly every reviewed response is an actual error.
\paragraph{Telehealth routing.}\newprose{This is an illustrative economic calculation, not a deployment evaluation. It assumes per-query costs of \$25 for physician review, \$10 for nurse review, and \$0.50 for an automated response. Under those assumptions and the stated routing thresholds, modeled cost falls from \$25{,}150 to \$11{,}654 per 1{,}006 queries, with 26 incorrect benchmark responses in the automated tier. The cost values and reviewer behavior are assumed, not measured.}
\subsection{Finance deployment}Consider an automated financial analysis tool where an LLM answers quantitative reasoning queries for investment analysts. Incorrect outputs could lead to costly trading errors, so a tiered system routes confident outputs directly, uncertain ones to a junior analyst, and low-confidence ones to a senior analyst for verification. We evaluate on 1{,}930 samples from quantitative reasoning benchmarks (GPQA, SimpleQA, LiveBench, BBEH, HLE, OmniMath, MathVista, MathVision, MathVerse), achieving 0.952 AUROC with a base accuracy of 43.7\%.
\paragraph{Analyst triage.}\newprose{Table~\ref{tab:finance_triage} reports error detection on the benchmark responses. Reviewing the bottom 50\% by calibrator score captures 83\% of observed errors. The 75\% ``junior analyst'' value is an assumed comparison point, not a measured human baseline.}
\begin{table}[t]\centering
\small
\caption{\newprose{\textbf{Confidence-ranked review concentrates errors on the finance benchmark subset.} This is an illustrative threshold analysis; no analyst baseline was measured.}}
\label{tab:finance_triage}
\begin{tabular}{@{}cccc@{}}\toprule
Review \% & Errors Caught & Errors Slipped & Catch Rate \\
\midrule10\% & 193 & 894 & 18\% \\
20\% & 383 & 704 & 35\% \\
30\% & 567 & 520 & 52\% \\
40\% & 744 & 343 & 68\% \\
50\% & 899 & 188 & 83\% \\
\bottomrule
\end{tabular}

\end{table}

\paragraph{Risk tiering.}The calibrator assigns 30\% of outputs to the GREEN tier (reliable) at 95.3\% accuracy, 11\% to YELLOW (verify) at 72.9\%, and 59\% to RED (unreliable) at 12.1\%. The false GREEN rate is 4.7\% (27 wrong outputs labeled reliable out of 571 in the GREEN tier).
\paragraph{Robo-advisor routing.}\newprose{This illustrative calculation assumes per-query costs of \$50 for senior review, \$20 for junior review, and \$0.50 for an automated response. Under the stated thresholds, modeled cost falls from \$96{,}500 to \$60{,}244 per 1{,}930 queries, with 27 incorrect benchmark responses in the automated tier. Neither the costs nor analyst performance were measured in this study.}
\section{Cross-domain transfer to text-to-image evaluation}
\label{app:finegrain}We test whether \textsc{Pinocchio} can transfer to a fundamentally different domain: detecting failure modes in text-to-image (T2I) generation. Using the FineGRAIN benchmark~\citep{hayes2025finegrain}, which evaluates T2I outputs across 27 failure modes with human-annotated labels, we frame T2I evaluation as visual QA (``Does this image accurately depict: \textit{[prompt]}?'') and evaluate on 3{,}750 samples across 5 T2I models.Without any T2I-specific training, the calibrator achieves AUROC 0.736 [0.719, 0.751]. With domain adaptation (leave-one-model-out CV on 5 T2I models), AUROC improves to 0.953 $\pm$ 0.028. This suggests \textsc{Pinocchio}'s learned correctness signals generalize beyond LLM text evaluation to visual quality assessment.
\section{Error analysis and failure cases}
\label{app:failures}Of 1{,}953 test examples, the calibrator makes 116 ``hard errors'': 48 confident-but-wrong ($p > 0.9$, incorrect; 2.5\% of test set) and 68 unconfident-but-right ($p < 0.1$, correct; 3.5\%). Table~\ref{tab:failure_benchmarks} shows the distribution across benchmarks.
\begin{table}[t]\centering
\small
\caption{\textbf{Hard errors are rare and spread thinly across benchmarks.} Failure case distribution by benchmark. CW = confident-but-wrong ($p > 0.9$, incorrect); UR = unconfident-but-right ($p < 0.1$, correct). Benchmarks sorted by total failure count.}
\label{tab:failure_benchmarks}
\begin{tabular}{@{}lrrr@{}}\toprule
Benchmark & CW & UR & Total \\
\midrule
RealWorldQA & 7 & 8 & 15 \\
SimpleQA & 1 & 9 & 10 \\
MM-Vet & 1 & 8 & 9 \\
VizWiz & 6 & 3 & 9 \\
ARC-AGI & 1 & 7 & 8 \\
MMStar & 5 & 3 & 8 \\
MathVista & 1 & 7 & 8 \\
CharXiv & 5 & 2 & 7 \\
BBEH & 1 & 4 & 5 \\
HLE & 0 & 5 & 5 \\
\midrule
All others (10 benchmarks) & 20 & 12 & 32 \\
\midrule
\textbf{Total} & \textbf{48} & \textbf{68} & \textbf{116} \\
\bottomrule
\end{tabular}

\end{table}

\paragraph{Systematic patterns.}The confident-but-wrong failures cluster in \emph{visual reasoning} benchmarks (RealWorldQA, VizWiz, MMStar, CharXiv) where the target model produces a plausible-sounding answer that is visually wrong. These are cases where the response text appears well-formed and confident, giving the calibrator insufficient signal that the answer is wrong.\newprose{The unconfident-but-right failures concentrate in SimpleQA, RealWorldQA, MM-Vet, ARC-AGI, MathVista, and HLE. Many of these errors are terse or use unconventional answer formats. This descriptive pattern does not establish that response length causes the low scores; the residualized analysis in Table~\ref{tab:length_analysis_v3} finds that length explains little of the overall discrimination signal.}
\section{Statistical significance}
\label{app:effect_size}
\begin{figure}[t]\centering
\includegraphics[width=0.85\textwidth]{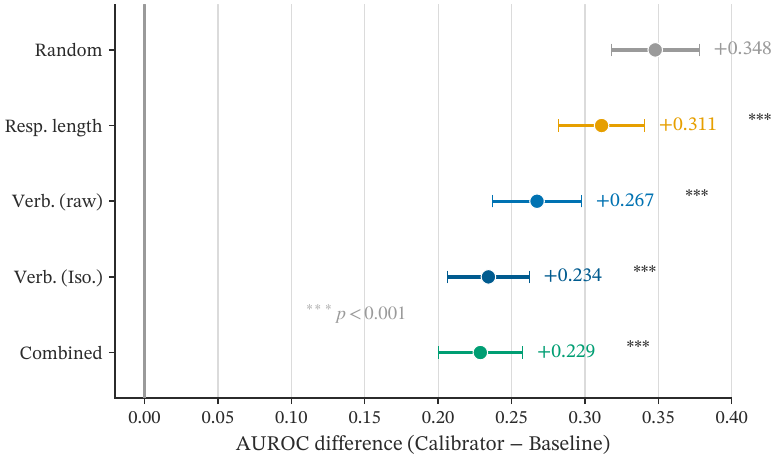}
\caption{\textbf{\textsc{Pinocchio} beats every full-set baseline by a wide, significant margin.} AUROC difference (calibrator minus baseline) with 95\% bootstrap CIs. The two proxy sampling baselines, evaluated on a different text-only subset, are excluded. All displayed differences are significant at $p < 0.001$.}
\label{fig:effect_size}
\end{figure}

We assess AUROC differences using paired comparisons only when methods are evaluated on the same examples. The full-set baselines use the held-out test set of 1{,}953 samples; the two proxy sampling baselines use the 1{,}376-example text-only subset and are excluded from the full-set significance statements below.\textbf{DeLong's test}~\citep{delong1988comparing} compares correlated AUROC curves directly. Comparisons between \textsc{Pinocchio} and each full-set baseline yield $p < 0.001$.\textbf{Bootstrap confidence intervals.} We compute 2{,}000 BCa bootstrap resamples of AUROC for the full-set methods. \textsc{Pinocchio}'s 95\% CI \purpleprose{[0.847, 0.879]} does not overlap with the full-set baseline intervals. The best full-set baseline is Combined at 0.649 [0.624, 0.673].\textbf{Effect sizes.} Figure~\ref{fig:effect_size} visualizes the AUROC difference between the calibrator and each full-set baseline with 95\% bootstrap CIs. The smallest gap is \purpleprose{+0.214} against Combined, and all displayed differences are significant at $p < 0.001$.

\section{Grading robustness and self-preference}
\label{app:self_preference}\newprose{Fifteen of the twenty benchmarks use exact-match or programmatic grading and therefore require no LLM judge. The remaining five use rubric-based grading with GPT-5-mini. A post-hoc label audit found grading edge cases in 18 of 12{,}972 labels (0.14\%), too few to affect the reported aggregate results. Table~\ref{tab:benchmarks_app} summarizes the benchmark mixture.}

\section{Benchmark descriptions}
\label{app:benchmarks}
\begin{table}[t]\centering
\footnotesize
\caption{\textbf{The benchmark mixture is broad and multimodal.} Benchmark mixture spanning seven domains and two modalities.}
\label{tab:benchmarks_app}
\begin{tabular}{@{}llc@{}}\toprule
\textbf{Domain} & \textbf{Benchmarks} & \textbf{Modality} \\
\midrule
Mathematics & OmniMath, MathVista, MathVerse, MathVision & T+V \\
Science & GPQA Diamond, ChemBench & T \\
Professional and coding & PRBench, ARC-AGI & T \\
General & SimpleQA, LiveBench, BBEH & T \\
Multimodal & MMMU, MMStar, MM-Vet, CharXiv, VizWiz & V \\
Spatial & RealWorldQA, HallusionBench & V \\
Frontier & HLE, HLE-Multimodal & T+V \\
\bottomrule
\end{tabular}

\end{table}

\subsection{Text benchmarks}
\begin{itemize}    \item \textbf{BBEH}~\citep{kazemi2025bbeh} (Big-Bench Extra Hard): 23 challenging reasoning tasks from BIG-Bench requiring multi-step inference, logical deduction, and causal reasoning. \vfiveprose{Strong} models achieve $\sim$50\% accuracy.    \item \textbf{GPQA Diamond}~\citep{rein2023gpqa}: 198 PhD-level multiple-choice questions in physics, chemistry, and biology, written by domain experts. Even expert humans achieve only $\sim$65\% accuracy.    \item \textbf{OmniMath}~\citep{gao2024omnimath}: Olympiad-level competition mathematics problems requiring symbolic reasoning and multi-step proofs.    \item \textbf{SimpleQA}~\citep{wei2024simpleqa}: Short-form factual questions with unambiguous, verifiable answers. Tests factual recall rather than reasoning.    \item \textbf{HLE}~\citep{phan2025hle} (Humanity's Last Exam): Expert-level questions across diverse academic domains, designed to be at the frontier of model capabilities.    \item \textbf{LiveBench}~\citep{white2024livebench}: Continuously updated benchmark with fresh questions. Covers math, coding, reasoning, and data analysis.    \item \textbf{ChemBench}~\citep{mirza2024chembench}: Chemistry knowledge and reasoning spanning organic, inorganic, and physical chemistry.    \item \textbf{PRBench}~\citep{akyurek2025prbench}: Professional reasoning benchmark with expert-written rubrics spanning legal and finance domains. Uses LLM-as-judge grading.    \item \textbf{ARC-AGI}~\citep{chollet2019measure}: Abstraction and Reasoning Corpus requiring novel pattern completion on grid-based visual tasks. Tests generalization to unseen transformation rules.
\end{itemize}

\subsection{Vision-language benchmarks}
\begin{itemize}    \item \textbf{MMMU}~\citep{yue2024mmmu}: Multimodal questions requiring college-level knowledge across 30 subjects including art, science, engineering, and medicine. Images include diagrams, charts, and photographs.    \item \textbf{MMStar}~\citep{chen2024mmstar}: Vision-indispensable questions specifically designed so that text-only models cannot solve them, ensuring genuine visual reasoning is required.    \item \textbf{CharXiv}~\citep{wang2024charxiv}: Questions about scientific figures and charts extracted from arXiv papers, requiring chart comprehension and numerical reasoning.    \item \textbf{HallusionBench}~\citep{guan2023hallusionbench}: Diagnostic suite for visual hallucination and illusion, testing whether models fabricate visual details or fall for optical illusions.    \item \textbf{MathVista}~\citep{lu2024mathvista}: Visual math reasoning combining geometric diagrams, statistical charts, and function plots with mathematical problems.    \item \textbf{MathVerse}~\citep{zhang2025mathverse}: Math problems where the visual diagram is necessary for correct interpretation; removing the image makes problems unsolvable.    \item \textbf{MathVision}~\citep{wang2024mathvision}: Competition-level math problems with diagram dependencies, sourced from mathematical olympiads.    \item \textbf{RealWorldQA}~\citep{xai2024realworldqa}: Spatial reasoning about real-world photographs, testing understanding of 3D scenes, object relationships, and physical properties.    \item \textbf{VizWiz}~\citep{gurari2018vizwiz}: Visual questions captured by blind users using smartphone cameras, including unanswerable queries due to image quality issues.    \item \textbf{MM-Vet}~\citep{yu2024mmvet}: Open-ended visual questions evaluating integrated capabilities including recognition, OCR, knowledge, spatial awareness, and language generation. Uses LLM-as-judge grading.
\end{itemize}

\section{Prompt templates}
\label{app:prompts}
\subsection{Calibrator prompt (combined format)}
\begin{verbatim}
Benchmark: {benchmark_name}
Source model: {source_model}
Question: {question}
Answer: {response}
Is the answer correct? (i) No (ii) Yes
\end{verbatim}
For VLM benchmarks, we prepend the image to the question in the model's native multimodal format. For text-only benchmarks, we use a 28$\times$28 gray placeholder image.
\subsection{Using the calibrator}
\paragraph{Using the calibrator.}\textsc{Pinocchio} is released as a pip package (\texttt{pip install pinocchio-uq}; source at \texttt{github.com/khayes95/pinocchio}, weights at \texttt{huggingface.co/KevinDavidHayes/pinocchio-0.8b}). \vfiveprose{The released checkpoint is a 0.8B model that performs on par with the 8B calibrator in the size ablation (Table~\ref{tab:model_size}).} It wraps an existing API call: the judge downloads its weights on first use and returns $\pcorrect$ from a single forward pass. The two additional lines referenced in the abstract are instantiating the judge and calling \texttt{score}.
\begin{verbatim}
from openai import OpenAI
from pinocchio import Pinocchio

client = OpenAI()
judge = Pinocchio()
messages = [{"role": "user", "content": question}]
response = client.chat.completions.create(
    model="gpt-5", messages=messages
)
p_correct = judge.score(response, messages=messages)
\end{verbatim}

\subsection{Verbalized confidence prompt}
\begin{verbatim}
You answered the following question:
Question: {question}
Your answer: {response}
How confident are you that your answer is correct?
Respond with only a number between 0 and 100,
where 0 means certainly wrong and 100 means
certainly correct.
Confidence:
\end{verbatim}

\section{Baseline implementation details}
\label{app:baselines}All post-hoc baselines use scikit-learn with a 50/50 train/test split (stratified by correctness label) for fitting. We did not tune hyperparameters; we use library defaults except where noted.
\begin{description}\item[Verbalized confidence (raw).]The target model's self-reported confidence, elicited via the prompt in Appendix~\ref{app:prompts}. Parsed as a float in $[0,1]$; used directly as $\hat{p}(\text{correct})$ with no post-processing.\item[Platt scaling (logistic regression).]A single-feature logistic regression fit on the verbalized confidence score: \texttt{LogisticRegression(C=1.0, solver=\textquotesingle lbfgs\textquotesingle, max\_iter=1000)}. The sigmoid output serves as the calibrated $\hat{p}(\text{correct})$. This is a standard two-parameter recalibration~\citep{platt1999probabilistic}.\item[Isotonic regression.]A non-parametric monotone recalibration of verbalized confidence: \texttt{IsotonicRegression(y\_min=0.01, y\_max=0.99, out\_of\_bounds=\textquotesingle clip\textquotesingle)}. Unlike Platt scaling, isotonic regression makes no parametric assumptions about the calibration curve and can correct non-sigmoid miscalibration. The number of segments is determined automatically by the pool-adjacent-violators algorithm (typically 10--30 pieces for our sample sizes).\item[Response length.]The feature is $\log(1 + t)$ where $t$ is the response token count, fit with the same logistic regression configuration as Platt scaling. The log transform handles the heavy-tailed length distribution (response lengths span $\sim$10 to $>$5{,}000 tokens). This baseline tests whether response verbosity alone predicts correctness.\item[Combined (verbalized + length).]A two-feature logistic regression on $[\text{verbalized\_confidence},\; \log(1+t)]$, using the same \texttt{LogisticRegression} configuration. Fit on a 50/50 train/test split to avoid overfitting the two-dimensional feature space.\item[LLM-as-judge (GPT-5-mini).]A zero-shot prompt asks GPT-5-mini to estimate $\pcorrect \in [0.0, 1.0]$ given only the question and response (no reference answer). We send requests asynchronously with up to 20 concurrent API calls. Response parsing: (1)~attempt direct float conversion; (2)~regex search for a decimal in $[0,1]$; (3)~if the response contains ``yes''/``correct'', assign 0.8; ``no''/``incorrect'', assign 0.2; (4)~otherwise assign 0.5. The fallback values are not optimized but are sufficient for AUROC evaluation: since AUROC depends only on rank ordering, any values satisfying $\text{no} < \text{fallback} < \text{yes}$ yield identical discrimination. The 0.5 fallback approximates the dataset base rate (53.4\% correct). We truncate questions and responses to 1{,}500 and 800 characters, respectively, matching the calibrator's input format.\item[Proxy semantic entropy.]We generate $N{=}5$ responses from Qwen3-VL-8B (temperature 0.7) for each question and compute the negative entropy of the semantic cluster distribution as a confidence score, following~\citet{kuhn2023semantic}. We call this a \emph{proxy} because the generations come from a different model than the target: since closed-source API models make repeated sampling prohibitively expensive, we use a small open-weight model as a stand-in. If the proxy model's uncertainty were informative about the target model's correctness, this would provide a cheap alternative to true semantic entropy. Evaluated on the 1{,}376 text-only test samples (VLM benchmarks excluded due to image input requirements).\item[Proxy self-consistency.]Using the same $N{=}5$ proxy generations from Qwen3-VL-8B, we compute the fraction that agree with the target model's response as a confidence score. We measure agreement by exact string match after normalization. As with proxy semantic entropy, this tests whether consistency among a proxy model's responses predicts correctness of the target model. Evaluated on the same 1{,}376 text-only samples.
\end{description}

\section{Faithful same-target sampling baselines}
\label{app:faithful_baselines}The proxy semantic-entropy and self-consistency baselines in Table~\ref{tab:main} sample a stand-in open model rather than the target. This is a low-cost workaround for closed APIs, but it does not satisfy the methods' same-target assumption. We therefore also sample an open target directly, LLaMA-3.1-8B-Instruct, on the hard test suite ($700$ questions) and compute each method from $N{=}10$ generations. We use the official implementations of semantic entropy~\citep{kuhn2023semantic}, the black-box estimators of \citet{lin2023generating}, SPUQ~\citep{gao2024spuq}, and the logit-free conformal method of \citet{su2024api}.
\begin{table}[t]\centering
\footnotesize
\caption{\textbf{Faithful same-target sampling baselines stay near random while \textsc{Pinocchio} leads.} Faithful same-target sampling baselines on LLaMA-3.1-8B ($N{=}10$, $700$ hard-suite questions), versus \textsc{Pinocchio} scored on the same responses. Paired $\Delta$ is the AUROC difference on matched questions, with uncertainty estimated from 1{,}000 paired bootstrap resamples.}
\label{tab:faithful_baselines}
\begin{tabular}{@{}lcc@{}}\toprule
Method & AUROC $\uparrow$ & Paired $\Delta$ vs.\ \textsc{Pinocchio} \\
\midrule
Semantic entropy (NLI) & 0.476 & $+0.335$ \\
Self-consistency (NLI) & 0.477 & $+0.334$ \\
\citet{lin2023generating} (best of 3) & 0.553 & $+0.258$ \\
SPUQ~\citep{gao2024spuq} & 0.482 & $+0.329$ \\
\citet{su2024api} (LofreeCP) & 0.574 & $+0.237$ \\
\midrule
\textsc{Pinocchio} (ours) & \textbf{0.811} & -- \\
\bottomrule
\end{tabular}

\end{table}
Every faithful baseline stays near random on the hard suite (Table~\ref{tab:faithful_baselines}), and the gap to \textsc{Pinocchio} is significant for all of them. Increasing the sample budget does not change this: same-target semantic entropy is $0.476 / 0.486 / 0.477$ at $N{=}10 / 20 / 40$. \newprose{The same pipeline produces higher AUROC on short-form QA benchmarks (TriviaQA 0.740, BoolQ 0.679, and NaturalQA 0.629), so the hard-suite result is not uniform across evaluation sets. We also obtain near-random semantic-entropy AUROC on DeepSeek-R1-Distill-Qwen-32B (0.479) and Claude Sonnet~4.6 (0.489 at 25.3\% accuracy). These additional targets show that the pattern is not confined to LLaMA-3.1-8B.}The result also replicates across lab-independent targets at larger scale (Table~\ref{tab:faithful_baselines_multi}): on Granite-4.1-30B ($n{=}1576$) and Devstral-2-24B ($n{=}1574$), every faithful baseline stays near random while \textsc{Pinocchio} reaches 0.806 and 0.836, paired $\Delta = +0.218$ and $+0.283$.
\begin{table}[t]\centering
\footnotesize
\caption{\textbf{\textsc{Pinocchio}'s advantage replicates across three lab-independent targets.} Faithful same-target sampling baselines on three lab-independent targets. Each baseline is run directly on $N{=}10$ generations from the target model itself on the hard suite, satisfying the methods' same-target assumption. Lin~(best) is the strongest of NumSemSets / Deg / EigV. The bottom row is the paired AUROC difference between \textsc{Pinocchio} and the strongest baseline for that target (1{,}000 paired bootstrap resamples).}
\label{tab:faithful_baselines_multi}
\begin{tabular}{@{}lccc@{}}\toprule
Method & \shortstack{LLaMA-3.1-8B\\
($n{=}700$)} & \shortstack{Granite-4.1-30B\\
($n{=}1576$)} & \shortstack{Devstral-2-24B\\
($n{=}1574$)} \\
\midrule
Semantic entropy (NLI)        & 0.476 & 0.558 & 0.514 \\
Self-consistency (NLI)        & 0.477 & 0.559 & 0.514 \\
\citet{lin2023generating} (best) & 0.553 & 0.534 & 0.520 \\
SPUQ~\citep{gao2024spuq}      & 0.482 & 0.588 & 0.553 \\
\citet{su2024api} (LofreeCP / conformal) & 0.574 & 0.521 & 0.497 \\
\midrule
\textsc{Pinocchio} (ours)     & \textbf{0.811} & \textbf{0.806} & \textbf{0.836} \\
\midrule
Paired $\Delta$ vs.\ best baseline & $+0.237$ & $+0.218$ & $+0.283$ \\
\bottomrule
\end{tabular}
\par\vspace{2pt}{\scriptsize Granite-4.1-30B and Devstral-2-24B are text-only; VLM benchmarks are excluded for those two targets.}
\end{table}
Table~\ref{tab:faithful_per_benchmark} breaks the hard suite down by benchmark for LLaMA-3.1-8B. \textsc{Pinocchio} wins in aggregate by a wide margin and on the hardest long-form benchmarks (\texttt{hle} 0.869, \texttt{livebench} 0.951), where the sampling methods fall to or below random. On a few lower-accuracy benchmarks whose answers are short enough that response diversity is informative, a sampling baseline is competitive or better (SPUQ on \texttt{omnimath} and \texttt{simpleqa}, the graph measure of \citet{lin2023generating} on \texttt{bbeh}), which is consistent with sampling-based uncertainty working precisely when consistency tracks correctness.
\begin{table}[t]\centering
\footnotesize
\caption{\textbf{\textsc{Pinocchio} wins overall and dominates on long-form benchmarks.} Per-benchmark AUROC on the hard suite for LLaMA-3.1-8B ($N{=}10$, $n{=}100$ per benchmark, text-only). SE and SC are NLI-based semantic entropy and self-consistency; Lin is the best of the three graph measures of \citet{lin2023generating}. Bold marks the best method on each benchmark. \textsc{Pinocchio} is scored black-box on the same responses.}
\label{tab:faithful_per_benchmark}
\begin{tabular}{@{}lrrcccccc@{}}\toprule
Benchmark & $n$ & Acc\% & SE & SC & Lin & SPUQ & LofreeCP & \textsc{Pinocchio} \\
\midrule
\texttt{bbeh}      & 100 & 3  & 0.352 & 0.347 & \textbf{0.644} & 0.478 & 0.629 & 0.641 \\
\texttt{chembench} & 100 & 43 & 0.493 & 0.517 & 0.468 & 0.568 & 0.579 & \textbf{0.583} \\
\texttt{gpqa}      & 100 & 29 & 0.516 & 0.517 & 0.534 & 0.556 & 0.506 & \textbf{0.598} \\
\texttt{hle}       & 100 & 14 & 0.532 & 0.532 & 0.503 & 0.549 & 0.630 & \textbf{0.869} \\
\texttt{livebench} & 100 & 53 & 0.404 & 0.405 & 0.682 & 0.422 & 0.541 & \textbf{0.951} \\
\texttt{omnimath}  & 100 & 6  & 0.582 & 0.582 & 0.709 & \textbf{0.876} & 0.766 & 0.783 \\
\texttt{simpleqa}  & 100 & 2  & 0.441 & 0.462 & 0.495 & \textbf{0.827} & 0.776 & 0.791 \\
\midrule
Overall            & 700 & 21 & 0.476 & 0.477 & 0.553 & 0.482 & 0.574 & \textbf{0.811} \\
\bottomrule
\end{tabular}

\end{table}
The per-benchmark SE/SC values also expose a second result: where sampling methods fail, they fail by \emph{inverting} below random, and the inversion is domain-specific rather than uniform. Strong inversion concentrates on benchmarks where the model commits to a single long-form wrong trajectory across all $N$ samples (\texttt{bbeh} 0.352, \texttt{livebench} 0.404); it weakens where some answer-space diversity survives (\texttt{chembench} 0.493, \texttt{gpqa} 0.516); and it disappears on \texttt{omnimath} (0.582), whose short numerical answers paraphrase cleanly enough for NLI clustering to work. A wiring error would invert every benchmark uniformly. The pattern instead follows the two mechanisms behind the collapse: confidently wrong trajectories that repeated sampling cannot escape, and NLI-based semantic clustering (DeBERTa-large-MNLI, trained on short sentence pairs) failing on multi-paragraph reasoning, where it neither merges paraphrases of the same correct answer nor separates distinct wrong answers of similar surface form.Even \emph{white-box} access to the target's token log-probabilities gives no usable correctness signal on the hard suite (Table~\ref{tab:logprob_baseline}): on LLaMA-3.1-8B, mean log-probability scores 0.413 AUROC (below chance), and no sequence-likelihood statistic exceeds verbalized confidence (0.610), while \textsc{Pinocchio}, scored black-box on the same responses, reaches 0.811.
\begin{table}[t]\centering
\footnotesize
\caption{\textbf{Even white-box log-probabilities carry no usable correctness signal on the hard suite.} White-box sequence-likelihood baselines on LLaMA-3.1-8B with full logit access, on the hard suite. Even with direct access to the target's token log-probabilities, sequence likelihood carries no usable correctness signal on hard reasoning, whereas \textsc{Pinocchio} (black-box, scored on the same responses) reaches $0.811$.}
\label{tab:logprob_baseline}
\begin{tabular}{@{}lc@{}}\toprule
Method & AUROC $\uparrow$ \\
\midrule
Mean log-prob (= $-$perplexity) & 0.413 \\
Min-token log-prob              & 0.447 \\
Sum log-prob                    & 0.521 \\
Token entropy                   & 0.520 \\
Verbalized confidence           & 0.610 \\
\midrule
\textsc{Pinocchio} (ours)       & \textbf{0.811} \\
\bottomrule
\end{tabular}

\end{table}

\section{Computational resources}
\label{app:compute}\newprose{Table~\ref{tab:compute} summarizes the compute used for training and evaluation.}
\begin{table}[t]\centering
\caption{\textbf{Training and evaluation are inexpensive.} Approximate training and evaluation times.}
\label{tab:compute}\small
\begin{tabular}{@{}ll@{}}\toprule
Task & Time \\
\midrule
\textsc{Pinocchio} training (\vfiveprose{32{,}419} examples) & 2--3 hours \\
Test set evaluation (1{,}953 examples) & $<$30 minutes \\
Use case scoring (1{,}953 examples) & $<$1 hour \\
\bottomrule
\end{tabular}

\end{table}

\paragraph{Training efficiency.}The use of LoRA adapters (rank 32) reduces trainable parameters to $<$1\% of the base model, enabling training on 4 GPUs with standard VRAM ($\geq$40GB per GPU for VLM training with images).
\paragraph{Inference cost.}At inference, \textsc{Pinocchio} requires a single forward pass through the 8B calibrator model per query ($\sim$0.1 seconds on a single GPU). This contrasts with sampling-based methods like semantic entropy~\citep{kuhn2023semantic} that require 5--10 forward passes through the \emph{target} model, representing a 5--10$\times$ reduction in cost per uncertainty estimate.
\paragraph{Smaller models.}\vfiveprose{A size ablation at fixed LoRA rank (Table~\ref{tab:model_size}) shows that calibrator size has little effect: held-out AUROC stays near 0.86 from 0.8B to 8B, and the released 0.8B model performs on par with the 8B calibrator.}
\begin{table}[t]\centering
\small
\caption{\vfiveprose{\textbf{Calibrator size has little effect.} Held-out AUROC, averaged over the four models whose responses train \textsc{Pinocchio}, across calibrator sizes at fixed LoRA rank. The 2B, 4B, and 8B calibrators share the Qwen3-VL backbone; the released 0.8B is a text-only Qwen3.5 model, evaluated on the text responses. Training-set size (Figure~\ref{fig:training_size}) has a much larger effect.}}
\label{tab:model_size}
\begin{tabular}{@{}llc@{}}\toprule
Calibrator & Params & Held-out AUROC $\uparrow$ \\
\midrule
\vfiveprose{Qwen3.5 (text-only, released)} & \vfiveprose{0.8B} & \vfiveprose{0.861} \\
\vfiveprose{Qwen3-VL} & \vfiveprose{2B} & \vfiveprose{0.857} \\
\vfiveprose{Qwen3-VL} & \vfiveprose{4B} & \vfiveprose{0.863} \\
\vfiveprose{Qwen3-VL (flagship)} & \vfiveprose{8B} & \vfiveprose{0.859} \\
\bottomrule
\end{tabular}
\end{table}

\begin{figure}[t]\centering
\includegraphics[width=0.7\textwidth]{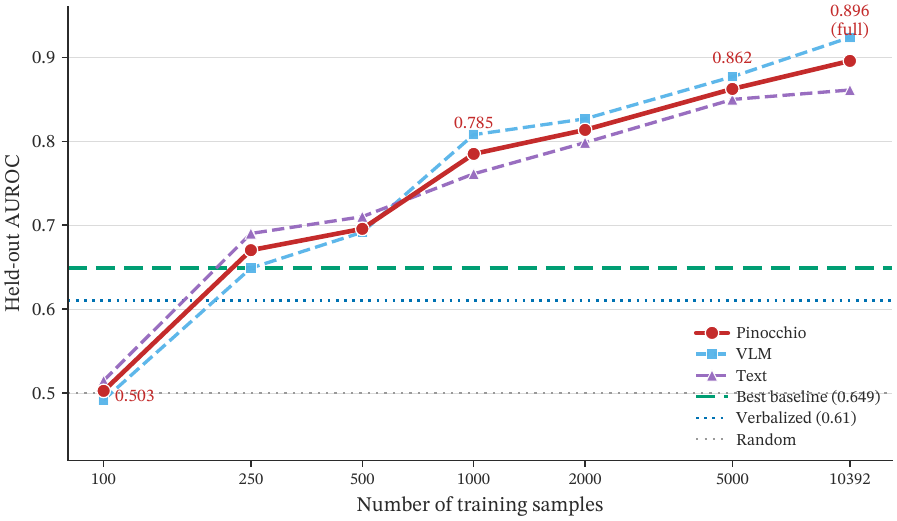}
\caption{\textbf{More training data helps more than a larger calibrator.} Training-set size. AUROC is measured on the ablation's held-out split, on which the full-data calibrator reaches 0.896; on the main test set it reaches the \purpleprose{0.863} reported throughout (Table~\ref{tab:main}). Calibrator parameter count has a much smaller effect (Table~\ref{tab:model_size}).}
\label{fig:training_size}
\end{figure}

\section{Additional main-body figures and tables}
\label{app:additional_figures}\newprose{Table~\ref{tab:lomo} reports the leave-one-model-out results.}
\begin{table}[t]\centering
\caption{\textbf{The calibrator transfers to a held-out source model with only a small drop.} Leave-one-model-out (LOMO) cross-model transfer. Each row trains on two source models and evaluates on the held-out third. $\Delta$ = difference from the full three-model calibrator (0.878).}
\label{tab:lomo}
\begin{tabular}{llcc}\toprule
Held-out model & Training models & AUROC & $\Delta$ \\
\midrule
GPT-5-mini & GPT-5.2 + Qwen3.5 & 0.877 & $-$0.001 \\
GPT-5.2 & GPT-5-mini + Qwen3.5 & 0.861 & $-$0.017 \\
Qwen3.5 & GPT-5-mini + GPT-5.2 & 0.790 & $-$0.088 \\
\midrule
\multicolumn{2}{l}{Mean (LOMO)} & 0.843 & $-$0.035 \\
\multicolumn{2}{l}{Best baseline (combined)} & 0.649 & -- \\
\bottomrule
\end{tabular}

\end{table}

\subsection{Training scale and input ablations}
\label{app:scaling}Performance improves most rapidly up to $N{=}1{,}000$ and then shows diminishing returns. \purpleprose{Removing benchmark and model-identity metadata reduces AUROC from 0.863 to 0.813. Removing only the model-identity tag gives 0.850 AUROC, with larger reductions on Qwen3.5 and VLM benchmarks.} The question-versus-response ablation is reported in Table~\ref{tab:knowledge_cues}.
\paragraph{Vision-language data as text augmentation.} Table~\ref{tab:vlm_augment} reports the text-budget sweep from Section~\ref{sec:per_benchmark}: calibrators trained on $N$ text examples, with and without the full vision-language training set added, all at matched configuration (LoRA rank 32, two epochs) and evaluated on the same held-out text benchmarks.
\begin{table}[t]\centering
\footnotesize
\caption{\textbf{Vision-language data improves text calibration when hard text data is scarce.} Text-benchmark AUROC for calibrators trained on $N$ text examples, with and without the vision-language training set added (matched configuration, same held-out text set). Gains are largest at small $N$ and taper as text data grows.}
\label{tab:vlm_augment}
\begin{tabular}{@{}cccc@{}}\toprule$N$ (text examples) & Text only & + vision-language & $\Delta$ \\
\midrule500    & 0.690 & \textbf{0.731} & $+0.041$ \\
1{,}000 & 0.730 & \textbf{0.789} & $+0.059$ \\
2{,}500 & 0.791 & \textbf{0.815} & $+0.025$ \\
\bottomrule
\end{tabular}

\end{table}

\paragraph{Response length.} The response content is load-bearing: truncating it degrades AUROC monotonically (Table~\ref{tab:truncation}). At a 200-character budget AUROC falls to \purpleprose{0.790}, and beyond the 800-character canonical setting it continues to rise slightly, so more of the response helps.
\begin{table}[t]\centering
\footnotesize
\caption{\textbf{Longer response context monotonically improves AUROC.} Response-truncation ablation. AUROC as a function of the response character budget, relative to the 800-character canonical setting.}
\label{tab:truncation}
\begin{tabular}{@{}lcc@{}}\toprule
Response budget & AUROC & $\Delta$ vs canonical \\
\midrule$r{=}200$            & \purpleprose{0.7898} & \purpleprose{$-0.074$} \\
$r{=}400$            & \purpleprose{0.8357} & \purpleprose{$-0.028$} \\
$r{=}800$ (canonical) & \purpleprose{0.8633} & -- \\
$r{=}1600$           & \purpleprose{0.8691} & \purpleprose{$+0.006$} \\
$r{=}3000$           & \purpleprose{0.8691} & \purpleprose{$+0.006$} \\
\bottomrule
\end{tabular}

\end{table}

\subsection{Per-model recalibration}
\label{app:recalibration}For each target model, we fit Platt scaling and isotonic regression on 100 labeled examples and evaluate on the remainder. The procedure changes the mapping from scores to probabilities without retraining the calibrator. \newprose{Table~\ref{tab:isotonic} reports the matched set results.}
\begin{table}[t]\centering
\footnotesize
\caption{\newprose{\textbf{Platt scaling preserves AUROC while cutting ECE.} Per-model recalibration on the matched ablation set with a 100-sample fit set. Aggregate is the unweighted mean across models.}}
\label{tab:isotonic}
\begin{tabular}{@{}lccccccc@{}}\toprule& \multicolumn{2}{c}{\textbf{Raw}} & \multicolumn{2}{c}{\textbf{Platt}} & \multicolumn{2}{c}{\textbf{Isotonic}} \\
\cmidrule(lr){2-3} \cmidrule(lr){4-5} \cmidrule(lr){6-7}\textbf{Model} & AUROC & ECE & AUROC & ECE & AUROC & ECE & $N_{\text{eval}}$ \\
\midrule
GPT-5-mini    & 0.878 & 0.106 & 0.878 & 0.067 & 0.865 & 0.076 & 551 \\
GPT-5.2       & 0.875 & 0.078 & 0.875 & 0.062 & 0.870 & 0.050 & 624 \\
Qwen3.5       & 0.869 & 0.103 & 0.869 & 0.064 & 0.864 & 0.064 & 473 \\
\midrule
\textbf{Aggregate} & \textbf{0.874} & \textbf{0.096} & \textbf{0.874} & \textbf{0.064} & \textbf{0.866} & \textbf{0.063} & N/A \\
\bottomrule
\end{tabular}

\end{table}
Platt scaling preserves AUROC because it is monotone. \newprose{On the matched ablation set, isotonic regression introduces ties while reducing aggregate ECE from 0.096 to 0.063.}
\paragraph{\vfiveprose{Held-out evaluation.}}
\label{app:frontier_trained}\vfiveprose{\textsc{Pinocchio} is trained jointly on seven LLMs that span a wide capability range. It achieves 0.862 AUROC on held-out responses from those models; Table~\ref{tab:frontier_flagship} \orangeprose{reports its per-model calibration.}} Recalibration also holds when we hold out whole \emph{domains} rather than models. We run four leave-one-domain-out folds, in each of which the calibrator excludes a domain group from training and then scores it, and we fit recalibration on about 100 labels from the held-out domain (Table~\ref{tab:domain_recal}). Recalibrating on those $\sim$100 labels cuts pooled ECE several-fold, from 0.271 off-the-shelf to 0.050 with Platt scaling, and adding the target's verbalized confidence as a feature does not consistently help beyond that. We note that AUROC on held-out domains is lower than on held-out models: domain holdout is the harder setting, and recalibration restores calibration but not the discrimination lost when an entire domain is unseen.
\begin{table}[t]\centering
\footnotesize
\caption{\textbf{Recalibration restores calibration even on held-out domains the calibrator never trained on.} Pooled ECE across four leave-one-domain-out folds. Off-the-shelf is the excluded-domain calibrator with no recalibration; the others fit on about 100 labels from the held-out domain. Verbal elicitation adds the target's self-reported confidence as a feature.}
\label{tab:domain_recal}
\begin{tabular}{@{}lc@{}}\toprule
Condition & Pooled ECE $\downarrow$ \\
\midrule
Off-the-shelf (no recalibration) & 0.271 \\
Domain scaling (Platt)           & 0.050 \\
Isotonic                         & 0.080 \\
Isotonic + verbal elicitation    & 0.065 \\
\bottomrule
\end{tabular}

\end{table}

\paragraph{Error analysis.}\newprose{On the matched 1{,}953-response analysis set, per-benchmark AUROC varies from 0.616 (HLE) to 0.995 (LiveBench), but benchmark accuracy does not explain this variation: Spearman $\rho=0.045$ ($p=0.85$). Two error patterns recur. On HLE and HLE-Multimodal, where target accuracy is 17\%, 50--63\% of the rare correct responses receive $\pcorrect<0.2$ and score separation is weak (Cohen's $d=0.30$--$0.54$). On PRBench, ChemBench, and GPQA, 17--22\% of incorrect answers receive $\pcorrect>0.8$. LiveBench, MathVista, and OmniMath instead exceed 0.929 AUROC. These observations identify where errors occur, but do not by themselves determine whether domain knowledge, response structure, or another factor causes the differences.}